%% file: Review.tex
\documentclass[twoside,11pt]{article}
\usepackage{jair, theapa, rawfonts}

\jairheading{63}{2018}{789-848}{04/18}{12/18}
\ShortHeadings{State-Space Abstractions for Probabilistic Inference}
{L\"udtke, Schr\"oder, Kr\"uger, Bader, \& Kirste}
\firstpageno{789}

\usepackage{amsmath,amsfonts,hyphenat,booktabs,graphicx,setspace, framed}
\usepackage[utf8]{inputenc}
\usepackage{listings}
\usepackage{subcaption}
\usepackage{tabularx}
\usepackage[autostyle=true]{csquotes}
\usepackage{amssymb}
\usepackage{longtable}
\usepackage{tabularx}
\usepackage{stmaryrd}
\usepackage{float}
\usepackage{amsthm}
\usepackage{xcolor}
\usepackage{paralist}
\usepackage{thmtools}
\usepackage{lipsum}

\theoremstyle{definition}

\usepackage{footnote}
\makesavenoteenv{tabular}

\usepackage{spverbatim}

\newcommand{\new}[1]{#1}

\newcommand\mymid{\!\mid\!}

\newcommand*\rot{\rotatebox{90}}

\declaretheoremstyle[
  bodyfont=\normalfont,
  spaceabove=1em plus 0.75em minus 0.25em,
  prefoothook=\hfill $\circ$,
  spacebelow=1em plus 0.75em minus 0.25em,
]{exmpstyle}

\declaretheorem[
  style=exmpstyle,
  title=Example,
  refname={example,examples},
  Refname={Example,Examples}
]{example}

\begin{document}

\title{State-Space Abstractions for Probabilistic Inference:\\ A Systematic Review}

\author{\name Stefan Lüdtke \email stefan.luedtke2@uni-rostock.de \\
       \addr 
        Institute of Computer Science\\
         University of Rostock, Germany
         \AND
       \name Max Schröder \email max.schroeder@uni-rostock.de \\
       \name Frank Krüger \email frank.krueger@uni-rostock.de \\
       \addr 
        Institute of Communications Engineering\\
         University of Rostock, Germany
         \AND
       \name Sebastian Bader \email sebastian.bader@uni-rostock.de \\
       \name Thomas Kirste \email thomas.kirste@uni-rostock.de \\
       \addr 
        Institute of Computer Science\\
         University of Rostock, Germany }

\maketitle

\begin{abstract}
Tasks such as social network analysis, human behavior recognition, or modeling biochemical reactions, can be solved elegantly by using the probabilistic inference framework.
However, standard probabilistic inference algorithms work at a propositional level, and thus cannot capture the symmetries and redundancies that are present in these tasks.

Algorithms that exploit those symmetries have been devised in different research fields, for example by the lifted inference-, multiple object tracking-, and modeling and simulation-communities.
The common idea, that we call \emph{state space abstraction}, is to perform inference over compact representations of sets of symmetric states.
Although they are concerned with a similar topic, the relationship between these approaches has not been investigated systematically. 

This survey provides the following contributions. We perform a systematic literature review to outline the state of the art in probabilistic inference methods exploiting symmetries. From an initial set of more than 4,000 papers, we identify 116 relevant papers. 
Furthermore, we provide new high-level categories that classify the approaches, based on common properties of the approaches. The research areas underlying each of the categories are introduced concisely.
Researchers from different fields that are confronted with a state space explosion problem in a probabilistic system can use this classification to identify possible solutions.
 Finally, based on this conceptualization, we identify potentials for future research, as some relevant application domains are not addressed by current approaches. 

\end{abstract}

\input{Input/01_Introduction.tex}
\input{Input/02_Preliminaries.tex}
\input{Input/03_Methods.tex}

\input{Input/04_Results.tex}

\input{Input/05_Analysis.tex}

\input{Input/06_Future_Research.tex}

\section*{Acknowledgements}
We are grateful to the three anonymous reviewers for their extensive comments and suggestions, which vastly improved the quality of the paper.

\appendix
\input{Input/07_AppendixDomainLifted.tex}
\input{Input/07_AppendixB.tex}

\newpage
\input{Input/07_AppendixC.tex}

\input{Input/07_AppendixD.tex}

\bibliography{aaa-complete-results-theoretical,ref,Reviews,Excluded-complete,further-manual-citations}
\bibliographystyle{theapa}

\end{document}

%% file: Input/01_Introduction.tex

\section{Introduction}
\label{sec:intro}

Many real-world problems are inherently symmetric. For example, human behavior recognition from sensor data \cite{fox_bayesian_2003}, social network analysis \cite{singla_lifted_2008}, and models of biochemical reactions \cite{barbuti_maximally_2011} all have symmetric properties.
These application scenarios are also probabilistic: We do not have perfect knowledge about the state of the system, and the system can develop non-deterministically over time.
Performing probabilistic inference in these domains quickly leads to a combinatorial explosion, known as \emph{state space explosion} problem \cite{clarke_progress_2001}. To overcome this problem, probabilistic inference approaches that exploit symmetric properties of the system have been devised.
In this survey, we systematically review the literature on these approaches and develop a new conceptual model to classify the approaches.
Previous surveys on this topic \cite{kersting_lifted_2012,kimmig_lifted_2015} have focussed on a specific class of such algorithms, known as \emph{lifted inference}. In this review, we put more emphasis on inference in sequential processes (known as \emph{Bayesian filtering}, a method that is highly relevant for many different application domains), and consider algorithms devised in a number of different research fields, like control theory, modeling and simulation, and computer vision.

To give an intuition of the state space explosion problem, we give some initial examples that show how it manifests itself in different domains. 

\begin{example}[Friends and Smokers, \shortciteR{singla_lifted_2008}]
\label{example:smokers}
The relationship of smoking habits and lung cancer is modeled. People who smoke are more likely to develop lung cancer, and friends tend to have similar smoking habits.
We can model this problem as a Bayesian network with one random variable for the smoking probability of each person, one random variable for the cancer risk of each person, and one random variable for each pair of people that represents whether they are friends or not.
The number of random variables and the treewidth of the graphical model grows linearly with the number of people, and thus the inference time grows exponentially (as inference is NP-hard in the treewidth of the model).
\end{example}

\begin{example}[Office, \shortciteR{fox_bayesian_2003}]
\label{example:tracking}
Several people walk around in an office. The office is equipped with presence sensors that get activated when a person is nearby. The sensors do not identify the specific person that is near the sensor.
The task is to keep track of  the locations of each person.
An inference algorithm has to track an exponential number of possible situations (all possible permutations of observations to person identities).
\end{example}

\begin{example}[Biochemical Reaction, \shortciteR{barbuti_maximally_2011}]
\label{example:chemistry}
Biochemical reactions can involve many different reactants.
In each specific reaction, many instances of the same molecule can participate in that reaction.
A naive algorithm has to consider an exponential number of specific reactions (one for each combination of specific molecule instances) that can take place. 
\end{example}
In all of these cases, standard probabilistic inference algorithms are not suitable, due to the exponential growth in problem complexity.
However, we can exploit the symmetries underlying each of these problems: In Example \ref{example:smokers}, the probability of each person having cancer is the same, as long as we have the same information about each person. We can therefore reason over all people simultaneously, by only representing the probability of a generic person having cancer. 
  In Example \ref{example:tracking}, people are not \emph{identified}. Thus, all states that are only different in the assignment of names to people cannot be distinguished and can be grouped together.
 In Example \ref{example:chemistry}, it does not matter which specific molecule participates in the reaction, as the result of the reaction is the same.
In all of the examples, the general idea is to represent multiple concrete (or \emph{grounded}) states that are symmetrical by a single abstract state (also called \emph{lifted} state).
In this paper, we identify two types of symmetries, based on \emph{exchangeability} in state variables or on variables following the same \emph{parametric distribution}.
In the following, we call the procedure of grouping symmetrical states \emph{state space abstraction}. 
To perform inference efficiently, an inference algorithm must be able to reason directly with the abstract states, without resorting to grounded states.

This systematic review aims at giving an overview of different methods of state space abstractions for probabilistic models, and inference algorithms that exploit these abstractions. 
The format of a \emph{systematic} literature review has been chosen because state space abstractions have been considered in different research communities (e.g.\ probabilistic inference, \shortciteR<see>{kersting_lifted_2012}; control theory, \citeR<see>{nitti_relational_2014};
modeling and simulation, \citeR<see>{maus_rule-based_2011}; computer vision,  \citeR<see>{huang_fourier_2009}, etc.). 
A systematic review is the appropriate tool in this case, because it reduces the chance to miss out relevant contributions from different research areas.

The contribution of this paper is a novel structure of the research field that is based on an \emph{application-centric} classification of the approaches. 
That is, approaches in the same class can exploit symmetries in the same problem domain.

Recently, attempts have been made to formally structure the problem classes of lifted inference algorithms, by investigating which structures of a probabilistic model allow efficient (lifted) probabilistic inference \cite{jaeger_liftability_2012}.
For lifted inference algorithms, this classification is precise and robust -- we present this classification in Appendix \ref{sec:li-complexity-classes}. However, it does not address algorithms for models containing continuous distributions, or dynamic models.
 In contrast, our classification is much more coarse-grained and informal (all problem classes they consider fall in the same category in our classification), but it applies to a broader range of algorithms. 
 
By using this classification, for the first time, this review draws connections between previously distinct lines of research, like lifted inference, logical filtering, and multiset rewriting, and outlines the common idea shared by these approaches -- the use of \emph{state space abstractions}.
We hope that this structure helps researchers from different research fields that are confronted with a state space explosion in a probabilistic system to identify possible solutions.
Finally, we identify potential future research directions.

We proceed as follows.
In Section \ref{sec:preliminaries}, we introduce the basic concepts used in the rest of the paper. Section \ref{sec:properties} contains a description of the properties that are used to characterize the approaches.
In Section \ref{sec:methods}, we describe the systematic procedure we applied for retrieving, selecting and analyzing the relevant work. 
An empirical overview of the retrieved papers is presented in Section \ref{sec:results}.
Section \ref{sec:analysis} contains the analysis of the retrieved papers, regarding the criteria proposed in Section \ref{sec:properties}. This evaluation leads to a categorization of the approaches, regarding the problem class they are concerned with. Each of the resulting groups is described separately.
We conclude in Section \ref{sec:futurework}, by discussing the results of this review, and proposing future research directions.

%% file: Input/02_Preliminaries.tex
\section{Preliminaries}
\label{sec:preliminaries}
This chapter gives a brief overview over basic concepts used in the remainder of the paper.
It consists of two parts: Section \ref{subsection:probabilisticinference} and \ref{subsec:DBN} introduce the basic concepts and algorithms used in the context of probabilistic inference.
Sections \ref{subsec:liftedgraphicalmodels} and \ref{subsec:rao-blackwellization} introduce the two basic concepts for state space abstractions that are discussed in this review: Lifted graphical models and Rao-Blackwellization. Each state space abstraction approach that we will discuss is based on either of these two concepts.

\subsection{Graphical Models and Probabilistic Inference}
\label{subsection:probabilisticinference}
In this section, we introduce the basic concepts of probabilistic inference, and briefly present three algorithms that are the basis for the lifted inference algorithms discussed in Section \ref{subsec:lifted-inference}. For a more thorough introduction to graphical models, see the book by \citeA{koller_probabilistic_2009}.

\subsubsection{Graphical Models}
\begin{figure}[tb]
\centering
\begin{subfigure}{0.3\textwidth}
\includegraphics[scale=0.5]{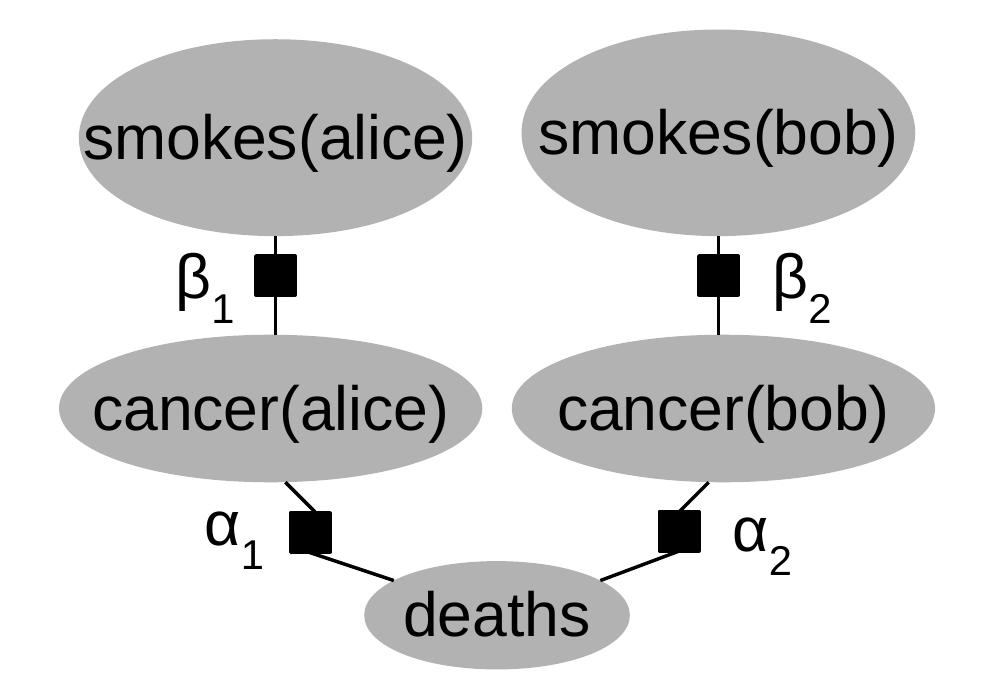}
\subcaption{Factor graph.}
\end{subfigure}
\begin{subfigure}{0.3\textwidth}
\centering
\begin{tabular}{lll}
\toprule
s(a) & c(a) & $\beta_1$\\
\midrule
0 & 0 & $\beta_1^{(00)}$ \\
0 & 1 & $\beta_1^{(01)}$ \\
1 & 0 & $\beta_1^{(10)}$ \\
1 & 1 & $\beta_1^{(11)}$\\
\bottomrule
\end{tabular}
\caption{Factor $\beta_1$. The values $\beta^{(xx)}$ are real numbers. The factor $\beta_2$ looks similar (see text).}
\label{subfig:betafactor}
\end{subfigure}
\hspace{0.1cm}
\begin{subfigure}{0.3\textwidth}
\centering
\begin{tabular}{lll}
\toprule
c(a)  & d & $\alpha_1$\\
\midrule
0 & 0 & $\alpha_1^{(00)}$ \\
0 & 1 & $\alpha_1^{(01)}$ \\
1 & 0 & $\alpha_1^{(10)}$ \\
1 & 1 & $\alpha_1^{(11)}$\\
\bottomrule
\end{tabular}
\caption{Factor $\alpha_1$. The values $\alpha^{(xx)}$ are real numbers. The factor $\alpha_2$ looks similar (see text).}
\label{subfig:alphafactor}
\end{subfigure}
\caption{Factor graph of Example \ref{ex:smokersbn} \shortcite<adapted from>{richardson_markov_2006}.}
\label{fig:smokers-grounded-models}
\end{figure}

Probabilistic graphical models are a way to compactly represent a joint probability distribution $P(X_1,\dots,X_n)$ that exhibits certain independence assumptions. 
They represent a joint probability distribution over multiple random variables (RVs) $X_1,\dots,X_n$ by decomposing the distribution $P(X_1,\dots,X_n)$ into a set of factors $F$. 
Each factor $\phi \in F$ maps a vector of RV assignments to non-negative real numbers, and the product of all factors describes the joint distribution (together with a normalization constant $Z$ ensuring that the total probability sums to one):
\begin{equation}
P(X_1=x_1,\dots,X_n=x_n)=Z^{-1}\prod_{\phi \in F} \phi (x_\phi)
\end{equation}
$x_\phi$ denotes the subset of values of RVs that is necessary to compute the factor $\phi$.
A factor of binary RVs is often represented as a table (for example, see Figure \ref{fig:smokers-grounded-models}).
A factor graph is a depiction of the relationship between factors and RVs. RVs are depicted by a circle, and factors by a box (see Figure \ref{fig:smokers-grounded-models}). Edges between factors and RVs mean that the RV is part of the factor.

Thus, graphical models provide a compact representation for probability distributions: Instead of representing a distribution over, for example, $n$ binary variables by a factor of size $2^n$ (a table with $2^n$ rows), the distribution is represented by a set of much smaller factors. This also makes reasoning about the distribution more efficient, as described later.

Bayesian networks and Markov networks
can be seen as special cases of factor graphs, where the factors are defined implicitly by the graph structure.
Bayesian networks are directed graphical models. The nodes represent RVs and an edge from a node $X$ to a node $Y$ means that the distribution of the RV $Y$ depends on the RV $X$.
Markov networks are undirected graphical models, where nodes represent RVs, and there is a factor for each maximal clique in the graph that takes the nodes of the clique as arguments. 

Consider the scenario introduced in Example \ref{example:smokers}.
We present a slightly adapted version of this scenario here (omitting the \emph{friends} relation for simplicity).
\begin{example}[Smokers]
\label{ex:smokersbn}
  Each person either smokes or does not smoke.
    For people who smoke, the chance of getting cancer is higher than for people who do not smoke.
Whether or not at least one person died last year depends on the number of people who have cancer.
\end{example}
For now, let us assume that only two people, Alice and Bob, exist. We can then model this scenario with the binary random variables $smokes(alice)$, $cancer(alice)$, $smokes(bob)$, $cancer(bob)$ and $death$\footnote{For readability, we use $c(a)$ and $c(b)$ instead of $cancer(a)$ and $cancer(b)$, $s(a)$ and $s(b)$ instead of $smokes(a)$ and $smokes(b)$, and $d$ instead of $death$.}.
The factor graph for this scenario can be seen in Figure \ref{fig:smokers-grounded-models}.

The factor graph describes a joint probability by multiplying all of the factors, for example:
\begin{equation}
\begin{split}
& P(s(a)=1,s(b)=1,c(a)=0,c(b)=0,d=0) \\
= & Z^{-1}\ \beta_1(s(a)=1,c(a)=0)\ \beta_2(s(b)=1,c(b)=0)\  \alpha_1(d=0,c(a)=0) \ \alpha_2(d=0,c(b)=0)
\end{split}
\label{eq:smokersfactor}
\end{equation}
Note that this example shows the need (and potential) for employing abstractions: We see that there is a certain redundancy in the model: The factors $\beta_1$ and $\beta_2$ as well as $\alpha_1$ and $\alpha_2$ are identical, when we exchange $s(a)$ and $c(a)$ for $s(b)$ and $c(b)$. If we want to add more people to the model, we need similar random variables and factors for each person. 
This behavior is the main motivation for employing state space abstractions: To be able to reason over these redundant variables as a group, ideally independently of the number of people (domain objects) involved.

\subsubsection{Inference Algorithms}
\label{subsubsec:inference-algs}
Given a graphical model, we can answer different questions. In our example, we may want to know the probability that Alice has cancer, or the expected number of deaths.
These questions fall into different categories: \emph{Conditional probability queries} $p(Q\mymid E{=}e)$, where the goal is to compute the conditional probability of some variables $Q$, given values of \emph{evidence} variables $E$, \emph{Maximum-a-posteriori (MAP) queries} $\text{MAP}(Q\mymid E{=}e) = \arg\max_q p(Q{=}q,E{=}e)$ that ask for the most likely joint assignment of variables, given values of evidence variables, and \emph{marginal MAP queries} $\text{MMAP}(S\mymid E{=}e)$ that ask for the most likely assignment of a subset $S \subset Q$ of variables, while the other variables $Q \setminus S$ are marginalized.

The process of calculating answers to these questions is called \emph{probabilistic inference}.
 Inference can always be performed by computing the complete joint distribution, and summing out (marginalizing) the variables we are not interested in. However, the reason for using graphical models in the first place was to avoid computing the complete joint distribution, so efficient inference algorithms avoid this. The remainder of this section will focus on conditional probability queries\footnote{MAP queries can be answered by adapting conditional probability inference algorithms (like variable elimination), or by specialized optimization algorithms. MMAP requires to calculate a marginal probability for each explored assignment of MAP variables, and thus in general is harder than the other query types. MMAP can be solved by search-based algorithms \cite{marinescu_pushing_2015}.}.

\paragraph*{Variable Elimination}
Variable elimination (VE) \shortcite{zhang_simple_1994} is an inference algorithm for conditional probability queries that operates on a factor graph. It eliminates the non-query and non-evidence variables one by one without computing the entire joint probability. A variable is eliminated by multiplying all factors that contain this variable, and then summing out (marginalizing) this variable. The performance depends on the order in which the variables are eliminated, and thus heuristics for good elimination orderings have been proposed \shortcite{darwiche_modeling_2009}.

\new{
\begin{example}
\label{ex:variable-elimination}
Consider the graphical model of Example \ref{ex:smokersbn} and the query $P(s(a),s(b),d{=}1)$\footnote{This query is the first step in answering the conditional probability query\\ $P(s(a),s(b)\mymid d)=P(s(a),s(b),d)/P(d)$.}. VE eliminates the non-query and non-evidence variables $c(a)$ and $c(b)$ one by one:
The RV $c(a)$ is eliminated by multiplying the factor $\alpha_1$ and $\beta_1$, resulting in a factor $f_0$ that has the following representation as a table (with 8 rows):
\begin{table}[H] 
\centering
\begin{tabular}{llll}
\toprule
s(a) & c(a)  & d & $f_0$\\
\midrule
0& 0 & 0 & $\beta_1^{(00)}\, \alpha_1^{(00)} $\\
0& 0 & 1 & $\beta_1^{(00)}\, \alpha_1^{(01)} $\\
\vdots & \vdots &  \vdots & \vdots \\
\bottomrule
\end{tabular}
\end{table}
The RV $c(a)$ is summed out of $f_0$, resulting in a factor 
\begin{equation*}
f_1(s(a),d)=\sum_v f_0(s(a),c(a){=}v,d) = \sum_v \beta_1(s(a),c(a){=}v)\, \alpha_1(c(a){=}v,d)
\end{equation*}
that is represented by the following table:
\begin{table}[H] 
\centering
\begin{tabular}{lll}
\toprule
s(a) & d & $f_1$\\
\midrule
0  & 0 & $\beta_1^{(00)}\, \alpha_1^{(00)} + \beta_1^{(01)}\, \alpha_1^{(10)}$ \\
0  & 1 & $\beta_1^{(00)}\, \alpha_1^{(01)} + \beta_1^{(01)}\, \alpha_1^{(11)}$ \\
\vdots  & \vdots  & \vdots \\
\bottomrule
\end{tabular}
\end{table}
 Thus, the distribution $P(s(a),s(b),c(b),d)$ can be represented by the factors $\alpha_2$, $\beta_2$ and $f_1$ as follows:
\begin{equation*}
P(s(a),s(b),c(b),d)=Z^{-1}\, f_1(s(a),d)\, \beta_2(s(b),c(b)) \, \alpha_2(c(b), d))
\end{equation*}
Afterwards, the same procedure is performed for $c(b)$: $\alpha_2$ and $\beta_2$ are multiplied, $c(b)$ is marginalized, the result is multiplied with $f_1$. The result directly represents the distribution of the above query.
\end{example}
In this example, the computations for eliminating $c(a)$ and $c(b)$ are similar, which hints to the possibility of performing the elimination more efficiently, as shown in Section \ref{subsec:lifted-inference}. 
}

\paragraph*{Recursive Conditioning}
\new{
Recursive conditioning (RC) \shortcite{darwiche_recursive_2001} is the search-based variant of VE. Instead of summing out RVs, it branches on the value of RVs. Once all information to evaluate a factor are present, it is evaluated directly, and the values of all branches are combined appropriately.
The presentation of RC given here is based on the description of \citeA{de_raedt_statistical_2016}.
}

\new{
Given a partially instantiated factor graph, the following cases are distinguished: (i) 
If there is a factor that can be evaluated, i.e.\ all RVs of this factor are instantiated, then it is evaluated, and RC is called on the remaining factor graph. The result of the factor evaluation and the RC call are multiplied. (ii) Otherwise, an RV is selected to branch on, RC is called recursively for each possible value of the RV, and the results of all recursive calls are summed. 
Furthermore, caching can be used to avoid repeated evaluation of the same expression, and disconnected components can be treated independently.
}

\begin{example}
\new{
Consider the same problem as in Example \ref{ex:variable-elimination}, i.e.\ the graphical model of Example \ref{ex:smokersbn} and the query $P(s(a),s(b),d{=}1)$.
RC starts with only $d=1$ instantiated, i.e.\ no factor can be evaluated. 
The algorithm selects $c(a)$ for branching, leading to the two branches $b_1$ where $\{d=1,c(a)=0\}$ and $b_2$ where $\{d=1,c(a)=1\}$. In both cases, the factor $\alpha_1$ can be evaluated, and the algorithm is called with the remaining factor graph. 
In the following, the algorithm branches on the other RVs $c(b)$, $s(a)$ and $s(b)$. The factor evaluations in each branch are multiplied, and the results of each branch are summed.
}
\end{example}

\paragraph*{Belief Propagation}
Belief propagation (BP) \cite{pearl_probabilistic_1988} is a message-passing inference algorithm, related to the forward-backward algorithm used in Hidden Markov Models. It is exact for acyclic factor graphs, and provides an approximate solution for factor graphs with cycles.  The idea is that each node (i.e.\ each RV node and each factor node) in a factor graph sends \emph{messages} to its neighbors, based on the messages it receives.

Let $x$ be an RV node (of the RV $x$) and $f$ be a factor node (of the factor $f$). 
Messages are passed either from an RV node to a factor node ($\mu_{x\rightarrow f}$) or from a factor node to an RV node ($\mu_{f\rightarrow x}$). The messages are partial functions with domain $\text{dom}(x)$, i.e. vectors of length ${|}\text{dom}(x){|}$.
The intuition on the messages $\mu_{f \rightarrow x}(x_j)$ is that the values are proportional to how likely node $f$ ``thinks'' the RV corresponding to node $x$ is in the state $x_j$.

More specifically, the messages are calculated as follows:
The message sent from an RV node $x$ to a factor node $f$ is the multiplicative summary of the message it received:
\begin{equation*}
\mu_{x\rightarrow f} (x_i) = \prod_{f'\in n(x) \setminus \{f\}} \mu_{f' \rightarrow x} (x_i)
\end{equation*}
$n(x)$ denotes the set of neighboring nodes of $x$ in the factor graph.
The message sent from a factor node $f$ to an RV node $x$ is
\begin{equation*}
\mu_{f\rightarrow x} (x_i) = \sum_{\mathbf{y}} \left( f(x_i,\mathbf{y}) \prod_{x' \in n(f) \setminus \{x\}} \mu_{x'\rightarrow f} (\mathbf{y}) \right)
\end{equation*}
The summation is over all possible assignments $\mathbf{y} \in \{dom(x')\mymid x' \in n(f) \setminus \{x\}\}$ of RVs $x'$ that are neighbors of $f$.
All messages $\mu_{x\rightarrow f}$ are initially set to 1. Then, the messages are updated until convergence.
For \emph{acyclic} factor graphs, belief propagation converges after a message has been sent and received by each node. 
For factor graphs with cycles, multiple iterations of sending and receiving messages can be performed (called loopy belief propagation). Conditions for convergence of the algorithm have been investigated by \citeA{weiss_correctness_2000}.

\new{
\begin{example}
Consider the factor graph of Example \ref{ex:smokersbn}. 
Here, we will not show the complete belief propagation algorithm, but only show how some of the messages are calculated.
The message $\mu_{c(a) \rightarrow \alpha_1}(x_{c(a)})$ with $x_{c(a)} \in \{0,1\}$ is updated according to 
\begin{equation*}
\mu_{c(a) \rightarrow \alpha_1} (x_{c(a)}) = \prod_{f' \in n(c(a)) \setminus \alpha_1} \mu_{f' \rightarrow c(a)} (x_{c(a)}) = \mu_{\beta_1 \rightarrow c(a) } (x_{c(a)})
\end{equation*}
The message $\mu_{\alpha_1 \rightarrow c(a)}(x_{c(a)})$ is updated according to 
\begin{equation*}
\mu_{\alpha_1 \rightarrow c(a)}(x_{c(a)}) = \sum_{\substack{x_d \in \{0,1\}}} \alpha_1(d{=}x_d,c(a){=}x_{c(a)}) \  \mu_{d \rightarrow \alpha_1} (x_{d})
\end{equation*}
\end{example}
}

\subsection{Bayesian Filtering}
\label{subsec:DBN}
An important subclass of probabilistic inference algorithms considers inference in cases where a distribution changes over time. They can be subsumed under the framework of \emph{Bayesian filtering} (also called recursive Bayesian state estimation) \cite{sarkka_bayesian_2013}. 
 For example, consider the following extension of Example \ref{ex:smokersbn}:
 \begin{example}
 \label{example:dynamicsmokers}
Smoking does not cause cancer immediately, but can cause cancer in the future. 
Having cancer does not immediately lead to death, but can cause death in the future.
Also, people who smoke tend to stay smokers, i.e. the probability of a person being a smoker depends on the person being a smoker at the previous time step.
\end{example}

\begin{figure}[tb]
\centering
\begin{subfigure}{0.39\textwidth}
\centering
\includegraphics[scale=0.45]{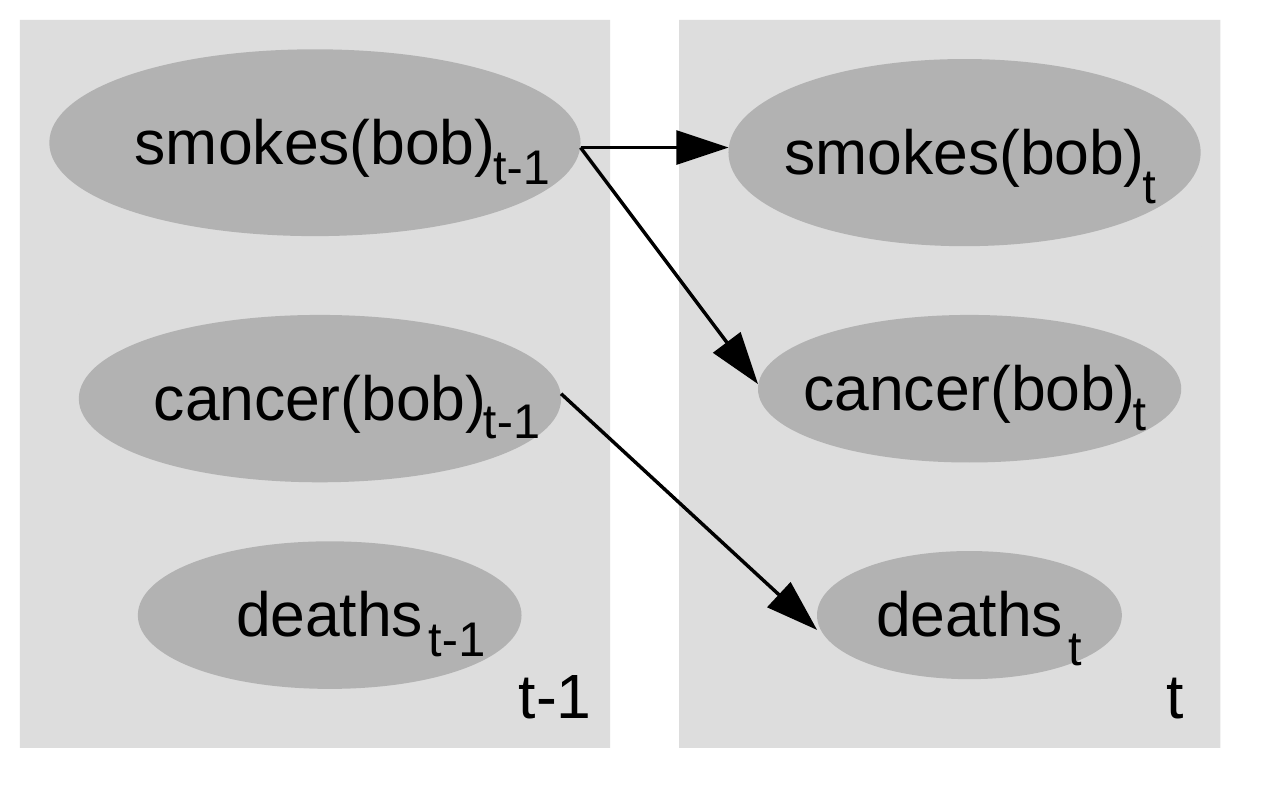}
\caption{Dynamic Bayesian Network}
\label{subfig:smokers-dynamic}
\end{subfigure}
\begin{subfigure}{0.6\textwidth}
\centering
\includegraphics[scale=0.45]{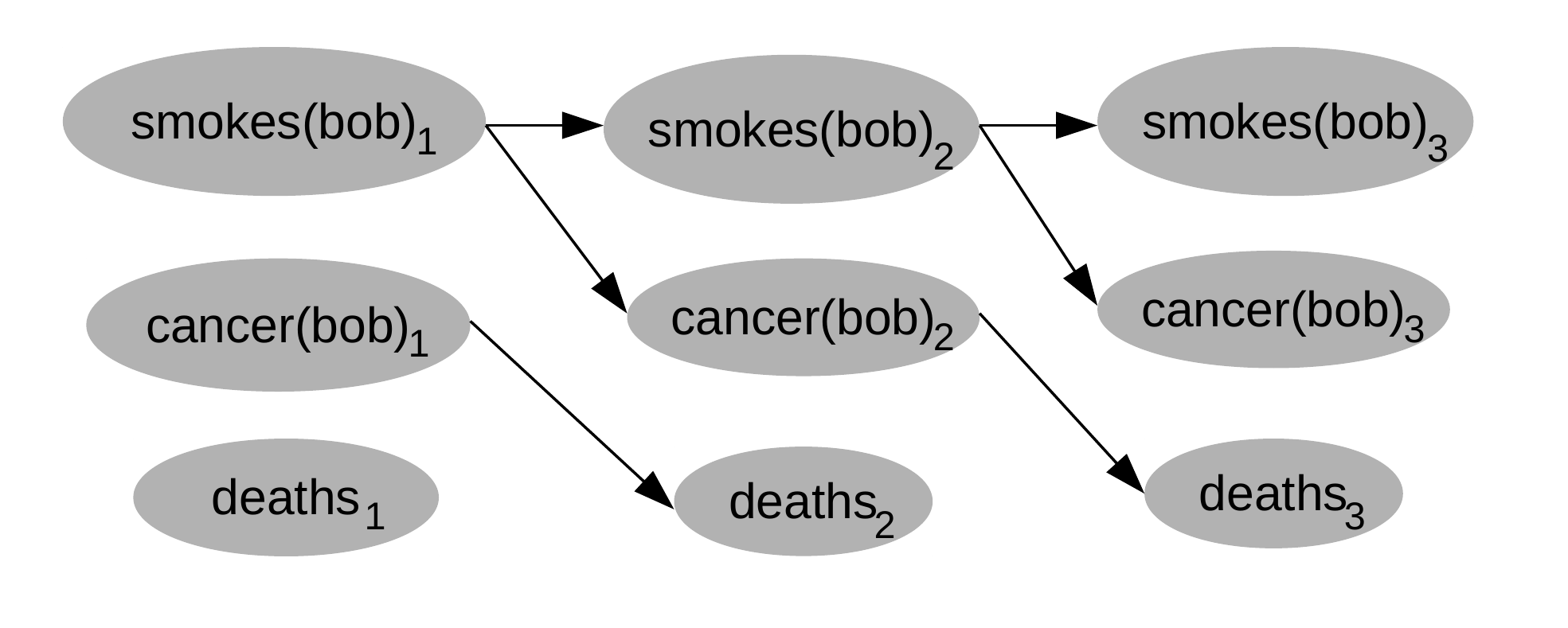}
\caption{Unrolled Bayesian Network}
\label{subfig:smokers-unrolled}
\end{subfigure}
\caption{Smokers domain with time dependencies (Example \ref{example:dynamicsmokers}). Light grey boxes indicate that the variables share the same time index.}
\label{fig:smokers-dbn}
\end{figure}

Such scenarios can be efficiently modeled by a dynamic Bayesian network (DBN). A DBN is essentially a Bayesian network with another dimension: There is a family of random variables indexed by time, and the value of each RV can depend on other RVs indexed by the same time, but also on RVs indexed by a previous time. That is, a DBN describes a stochastic process that has the Markov property.
The inference goal in a DBN is to estimate the state of some (not observed, or hidden) variables, given a sequence of observations of the other variables. This task is known as \emph{Bayesian filtering}.
In the example, we might get information about the number of deaths for each time step, and want to estimate the number of smokers per time step.

This task can be solved by viewing the DBN as standard graphical model (known as ``unrolling''), see Figure \ref{subfig:smokers-unrolled}.
Unrolling requires a finite observation sequence, and the sequence must be completely known to construct the unrolled network. 
However, for applications like sensor data processing, the observations sequence is of arbitrary length, and the observation sequence is not completely present at the beginning. Instead, the inference algorithm must be able to process the observations ``as they arrive'', without having access to ``later'' observations.

\begin{figure}[t]
\centering
\includegraphics[scale=0.5]{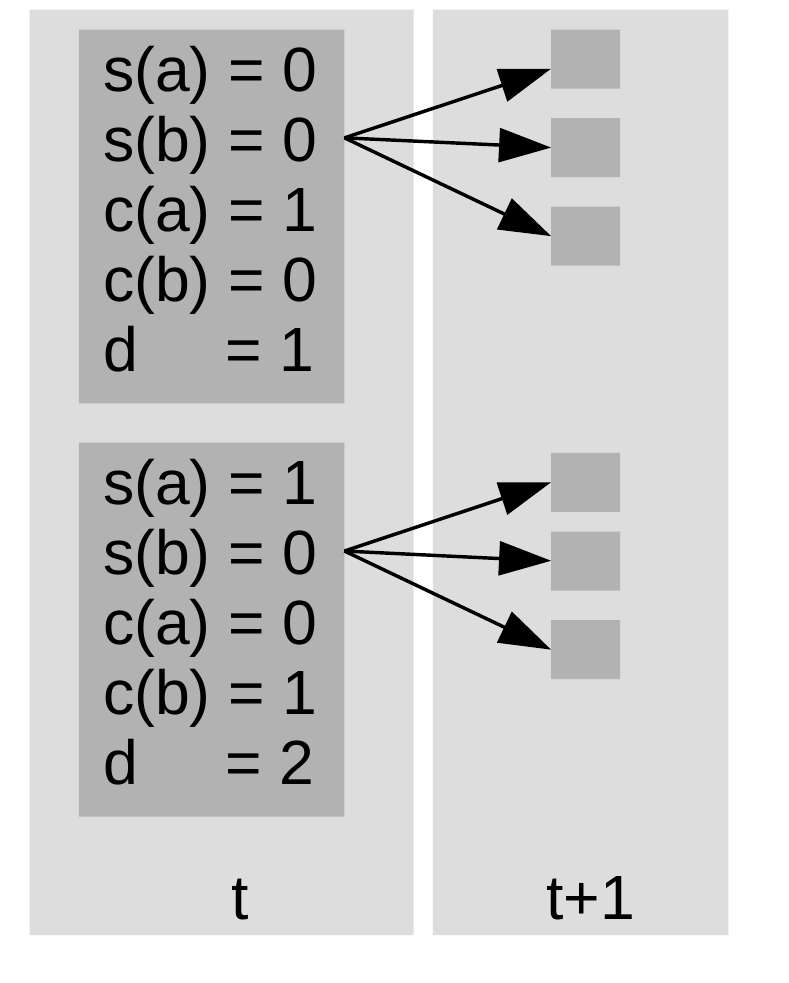}
\caption{Predict step of the particle filter for Example \ref{example:dynamicsmokers}. The example shows two particles at time $t$. Each particle has three successor states, leading to six particles at time $t+1$. The update step is not shown. Light grey boxes indicate the time index.}
\label{fig:particle-filter}
\end{figure}

Efficient algorithms for Bayesian filtering estimate the hidden state sequence $x_1,\dots,x_t$ recursively over time, given the observation sequence $y_1,\dots,y_t$.
To do so, the DBN is factored into a \emph{transition model} and an \emph{observation model}. 
The transition model $p(x_{t+1}\mymid x_{t})$  describes how the hidden state at time $t$ influences the hidden state at time $t+1$. 
The observation model $p(y_t\mymid x_t)$ describes how the hidden state at time $t$ influences the observation at the same time step.
The inference procedure is usually decomposed into two steps: In the \emph{prediction}, the state distribution for the next time step is calculated, based on the state distribution at the current time and the transition model, and by marginalizing over the current state: 
\begin{equation}
p(x_{t+1}\mymid y_{1:t})=\int p(x_{t}\mymid y_{1:t})\, p(x_{t+1}\mymid x_t)\, dx_t
\end{equation}
Afterwards, the predicted state is \emph{updated}, based on the observation:
\begin{equation}
p(x_{t+1}\mymid y_{1:t+1})=\frac{p(y_{t+1}\mymid x_{t+1})\, p(x_{t+1}\mymid y_{1:t})}{p(y_{t+1}\mymid y_{1:t})}
\end{equation}
Two well-known algorithms that implement this framework are the Kalman filter and the Hidden Markov Model. They can only be used for linear-gaussian models or models with finite state spaces, respectively.
In general, solving these equations exactly is infeasible.
A popular Monte-Carlo algorithm for Bayesian filtering is the particle filter \cite{doucet_sequential_2001}. 
The idea is to approximate the distribution $p(x_{1:t}\mymid y_{1:t})$ (the belief state) by a set of weighted samples.  The predict and update steps are performed on these particles. That is, a new set of particles is obtained by sampling from the transition distribution, conditioned on the current particles. Afterwards, each particle is updated according to the observation model. The algorithm is visualized in Figure \ref{fig:particle-filter}.

The state space explosion problem is also evident in many dynamic models: In Example \ref{example:dynamicsmokers}, the number of possible states per time step increases exponentially with the number of people.

\subsection{Lifted Graphical Models}
\label{subsec:liftedgraphicalmodels}
As discussed above, graphical models for situations that contain redundancies exhibit a symmetrical structure (cf. Example \ref{ex:smokersbn}). 
Lifted graphical models (also known as relational graphical models) 
provide a more compact syntactic representation for these cases. 
They provide a basis for lifted inference algorithms that allow to perform inference directly on this compact syntactic representation, avoiding redundant computations. 
In the following, we will introduce \emph{parfactor graphs}, one of the most common lifted graphical model formalisms.

\begin{figure}[tb]
\centering
\begin{subfigure}{0.3\textwidth}
\includegraphics[scale=0.5]{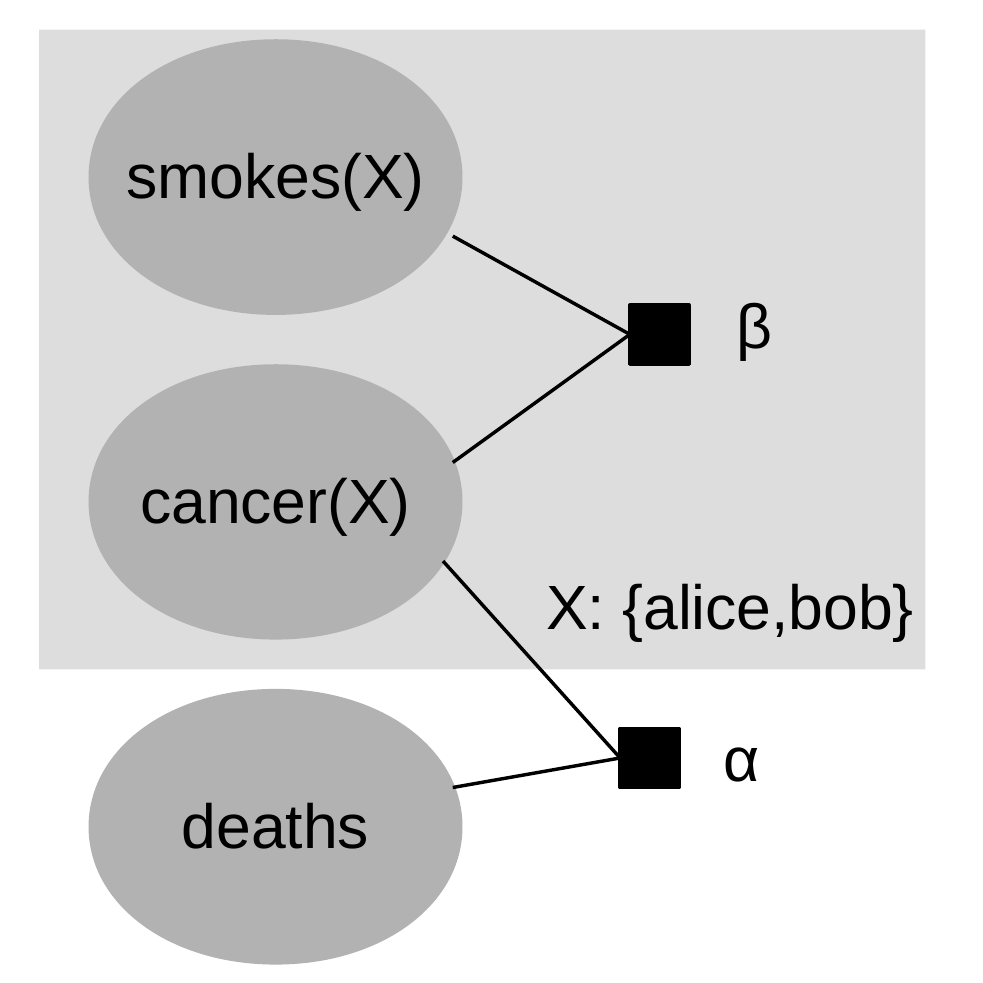}
\subcaption{Parfactor graph.}
\end{subfigure}
\begin{subfigure}{0.3\textwidth}
\centering
\begin{tabular}{lll}
\toprule
s(X) & c(X) & $\beta$\\
\midrule
0 & 0 & $\beta^{(00)}$ \\
0 & 1 & $\beta^{(01)}$ \\
1 & 0 & $\beta^{(10)}$ \\
1 & 1 & $\beta^{(11)}$\\
\bottomrule
\end{tabular}
\caption{Parfactor $\beta$.}
\label{subfig:betaparfactor}
\end{subfigure}
\begin{subfigure}{0.3\textwidth}
\centering
\begin{tabular}{lll}
\toprule
$c(X)$ & d & $\alpha$\\
\midrule
0 & 0  & $\alpha^{(00)}$ \\
0 & 1  & $\alpha^{(01)}$ \\
1 & 0  & $\alpha^{(10)}$ \\
1 & 1  & $\alpha^{(11)}$\\
\bottomrule
\end{tabular}
\caption{Parfactor $\alpha$.}
\label{subfig:alphaparfactor}
\end{subfigure}
\caption{Parfactor graph for Example \ref{ex:smokersbn}, using par-RVs and plate notation \cite{buntine1994operations}.}
\label{fig:smokers-parfactor}
\end{figure}

Parfactor graphs have been introduced by \shortciteA{poole_first-order_2003}.
They are motivated by the redundancies that can occur in factor graphs.
The idea of parfactor graphs is to represent the redundant factors (e.g.\ the factors $\beta_1$ and $\beta_2$ in Example \ref{ex:smokersbn}) only once.

Parfactor graphs achieve this by extending factor graphs by a first-order language. Factor graphs are related to parfactor graphs in the same way that propositional logic is related to first-order logic.
A \emph{parametric random variable} (par-RV) represents a set of random variables, one for each assignment of constants the parameters. The domain of each parameter is called \emph{population} (i.e.\ a set of individuals). For example, if $X$ is a parameter with the domain $\{a,b\}$, then $s(X)$ is a par-RV, and the parameter assignments $s(a)$ and $s(b)$ both represent a random variable. We call these RVs the \emph{groundings} of the par-RV.

A parametric factor, or \emph{parfactor}, is a function that maps par-RV assignments to the non-negative reals. For discrete RVs, the parfactor can be represented as a table. For example, the parfactor $\beta$ of Example \ref{ex:smokersbn} is shown in Figure \ref{subfig:betaparfactor}. Note that the factor is not indexed by the parameters of the par-RVs, i.e.\ the parfactor does not depend on the specific parameter assignments of the par-RVs.
A parfactor represents a set of factors, one for each grounding of the par-RVs.
 For example, the parfactor $\beta(s(X),c(X))$ represents the two factors $\beta_1(s(a),c(a))$ and $\beta_2(s(b),c(b))$. These factors are called the \emph{groundings} of the parfactor.

A set of par-RVs and parfactors can be represented by a \emph{parfactor graph}. The parfactor graph for Example \ref{ex:smokersbn} is shown in Figure \ref{fig:smokers-parfactor} (using plate notation,  \shortciteR{buntine1994operations}).
A parfactor graph defines a joint probability distribution as the normalized product of all groundings of the parfactors.
However, the joint distribution can also be calculated directly, without grounding all parfactors: Parfactors with the same truth assignment of variables need to be evaluated only once, raised to the power of the number of corresponding factors. For example, the probability calculated in Equation \ref{eq:smokersfactor} can be calculated as:
\newcommand*{\bigs}[1]{\vcenter{\hbox{\scalebox{1.2}{\ensuremath#1}}}}
\begin{align}
\begin{split}
& P\bigs{(}s(a){=}1,s(b){=}1,c(a){=}0,c(b){=}0,d{=}0\bigs{)}  \\
= & Z^{-1}\ \beta_1\bigs{(}s(a){=}1,c(a){=}0\bigs{)}\ \beta_2\bigs{(}s(b){=}1,c(b){=}0\bigs{)}\ \alpha_1\bigs{(}d{=}0,c(a){=}0\bigs{)} \ \alpha_2\bigs{(}d{=}0,c(b){=}0\bigs{)}\\
= & Z^{-1}  \prod_{X \in \{a,b\}} \beta{(}s(X){=}1,c(X){=}0\bigs{)}\ \alpha\bigs{(}d{=}0,c(X){=}0\bigs{)}\\
= & Z^{-1}\ \beta\bigs{(}s(X){=}1,c(X){=}0\bigs{)}^2\  \alpha\bigs{(}d{=}0,c(X){=}0\bigs{)}^2
\end{split}
\label{eq:smokersparfactor}
\end{align}
Compare this with Equation \ref{eq:smokersfactor}, where the factors $\beta_1$ and $\beta_2$ are evaluated and multiplied separately.
This example shows that inference operations can exploit the compact syntactic representation. Probabilistic inference algorithms that directly work on this representation are presented in Section \ref{subsec:lifted-inference}.

Multiple other lifted graphical model formalisms have been devised. 
A popular formalism are \emph{Markov logic networks} (MLNs) \cite{richardson_markov_2006}. MLNs are an extension of first-order logic with means to express uncertainty by assigning each first-order formula a weight that describes the tendency of the formula being violated. 
Other formalisms are based on paradigms like probabilistic logic programming \cite{kersting_1_2007,fierens_context-specific_2010}, or object orientation \cite{koller_object-oriented_1997,torti_reinforcing_2010-1}. 
A detailed description of representational formalisms is provided by \shortciteA{kimmig_lifted_2015}.
\new{
In general, a main difference between these formalisms is whether they are \emph{directed} or \emph{undirected}. Directed models can be interpreted in terms of conditional probabilities. The weights of undirected models cannot be interpreted locally, all weights together define the probabilistic model. 
In contrast to propositional graphical models, directed and undirected lifted models cannot be translated into each other in general.
Differences of the representation formalisms are discussed by  \shortciteA{de_raedt_statistical_2016}.
}

In this review, we focus on parfactor graphs, as they are easy to understand and 
allow a simple description of the exemplary lifted inference algorithms shown in Section \ref{subsec:lifted-inference} to illustrate the basic idea of lifted inference.

\subsection{Rao-Blackwellization}
\label{subsec:rao-blackwellization}
Apart from lifted graphical models, we consider a second type of state space abstraction in this review, called \emph{Rao-Blackwellization}. Lifted graphical models exploit the fact that multiple RVs are similar, i.e.\ symmetries between multiple RVs. Opposed to that, Rao-Blackwellization exploits the fact that the (conditional) distribution of several (often, but not necessarily continuous) RVs  follows a certain regular structure.
The idea is to represent such a distribution not explicitly (e.g.\ as a table of all possible values or a set of samples), but \emph{parametrically}.
 For example, consider a bivariate distribution $p(a,b) = p(a)\, p(b{|}a)$. Suppose that the conditional distribution $p(b{|}a)$ has some regular structure (e.g.\ it follows a normal distribution). 

For storing and manipulating this parametric function, the function needs to have a finite representation, like the string ``$\mathcal{N}(0,1)$''\footnote{If we only consider normal distributions, we could also represent it by a pair of reals. However, if we allow arbitrary parametric functions (that can have different numbers of parameters), a more flexible structure like a string is required.}. The \emph{semantics} of this \emph{syntactic structure} is the normal distribution with mean 0 and variance 1.

A well-known use of Rao-Blackwellization is the Rao-Blackwellized particle filter (RBPF) \cite{doucet_rao-blackwellised_2000}. In a RBPF, the state is decomposed, such that some RVs can be represented parametrically. The transition and observation model of the RBPF have to be able to maintain this representation appropriately, i.e.\ it must be possible to represent the posterior distribution (the distribution after performing one predict-update step) of these variables parametrically again.
This means that fewer particles are necessary to represent the belief state, because a distribution over fewer variables needs to be represented explicitly by samples. Thus, the belief state can be represented more compactly. 
The Kalman filter can be seen as the extreme case of a RBPF, where all variables are represented parametrically (by a normal distribution), and the transition model is linear.
Note that Rao-Blackwellization is orthogonal to lifted graphical models: Lifted graphical models represent graphical models with symmetrical variables compactly by grouping them, Rao-Blackwellization represents the distribution of a single or multiple variables compactly.

\new{
\begin{example}
\label{expl:smokers-number-deaths}
Suppose that we do not want to model whether or not a death occurred in Example \ref{ex:smokersbn}, but the \emph{number} $n_t$ of deaths, i.e.\ $n$ is an $\mathbb{N}$-valued RV, and we have a single factor $\alpha$ on all $c(X)$ RVs and the $n$ RV. Instead of representing the factor $\alpha$ explicitly by a table of exponential size, we can represent the number of deaths by a binomial distribution of the number $\#_P(c(P){=}1)$ of people with cancer: $n \sim binom(\#_P(c(P){=}1),p_d)$. This representation is much smaller (constant size in the number of $c(X)$ RVs). However, whenever the factor $\alpha$ needs to be manipulated (i.e.\ marginalizing RVs), this either has to be done on the parametric level (which may not be trivial), or the representation as a table has to be generated (which we try to avoid due to the exponential size of the table). 
\end{example}
In general, such a parametric representation is only possible for certain distributions, more specifically distributions that can be represented syntactically by a closed form mathematical expression.
}

\section{Properties of Inference Algorithms}
\label{sec:properties}

\begin{figure}[tb]
\centering
\begin{subfigure}{0.4\textwidth}
\centering
\includegraphics[scale=0.5]{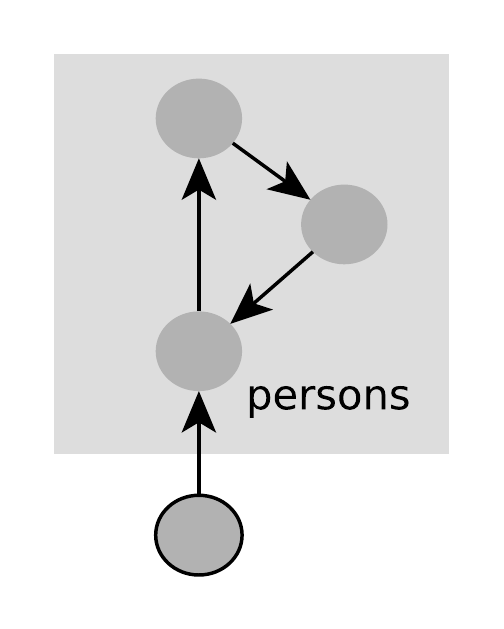}
\subcaption{Group Variables. Equivalent variables represented as a group (``lifted inference'').}
\end{subfigure}
\hspace{0.2cm}
\begin{subfigure}{0.4\textwidth}
\vspace{1cm}
\centering
\includegraphics[scale=0.5]{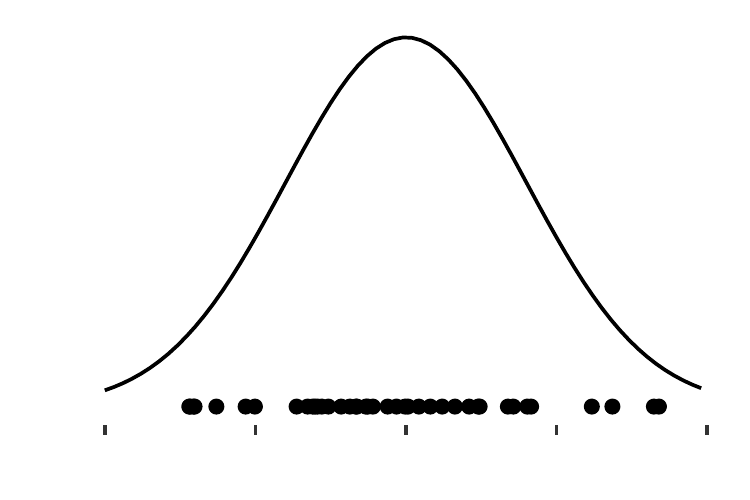}
\subcaption{Parameterization. The distribution of some state variables is represented parametrically, instead of explicitly by samples.}
\end{subfigure}
\begin{subfigure}{0.45\textwidth}
\centering
\includegraphics[scale=0.45]{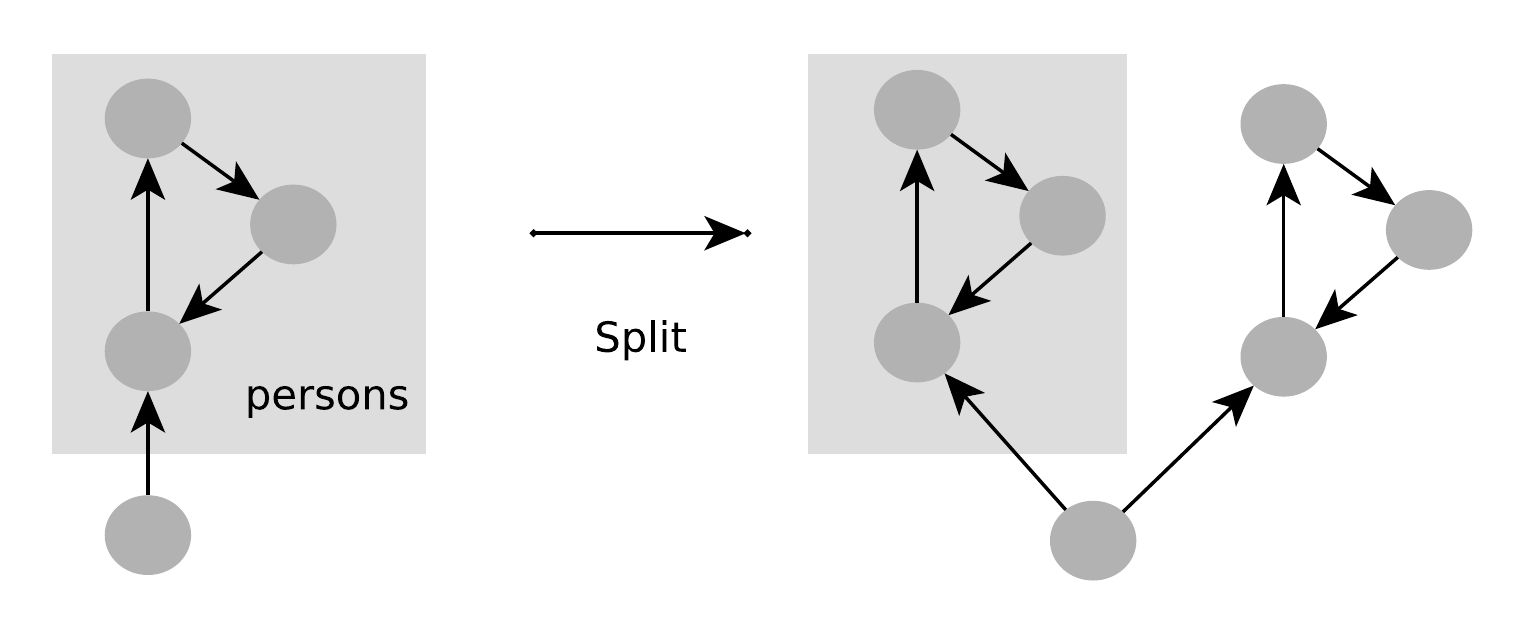}
\subcaption{Splitting. An operation that obtains a more specific representation.}
\end{subfigure}
\hspace{0.2cm}
\begin{subfigure}{0.45\textwidth}
\centering
\includegraphics[scale=0.45]{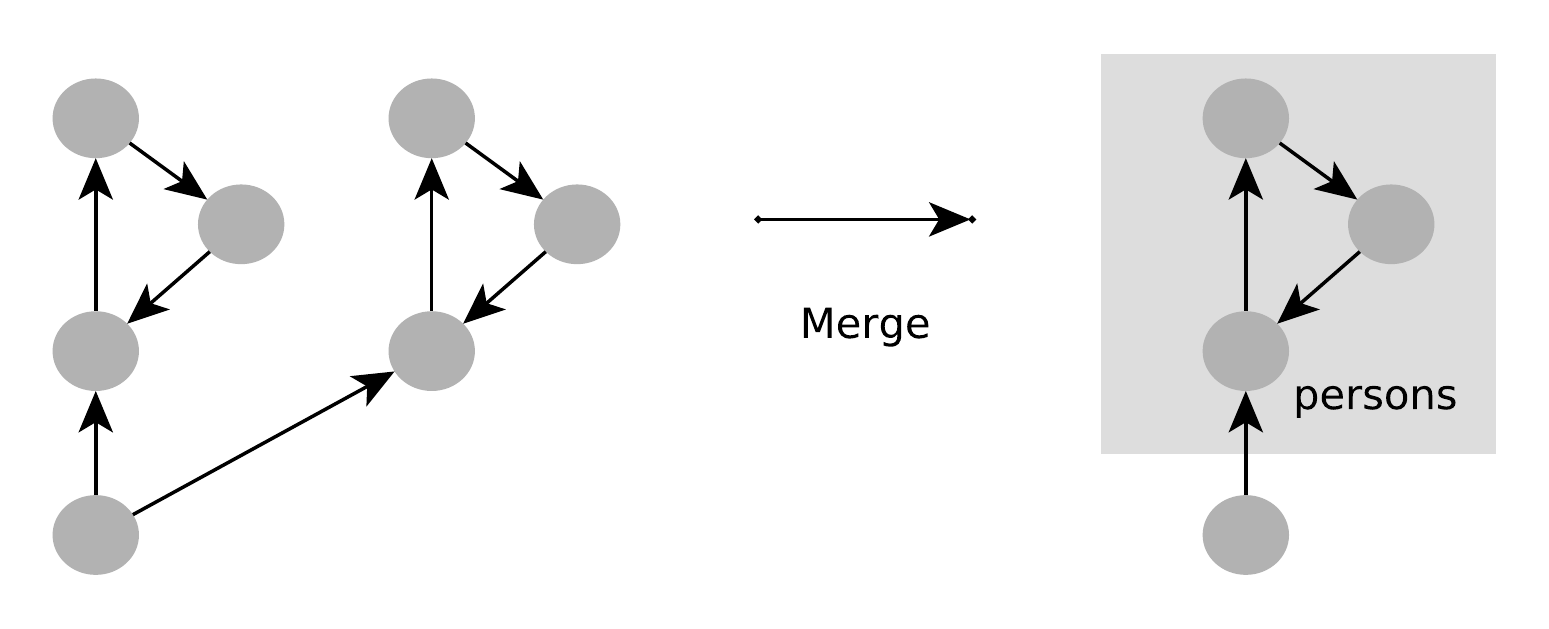}
\subcaption{Merging. An operation that obtains a more abstract representation.}
\end{subfigure}
\begin{subfigure}{0.4\textwidth}
\centering
\includegraphics[scale=0.5]{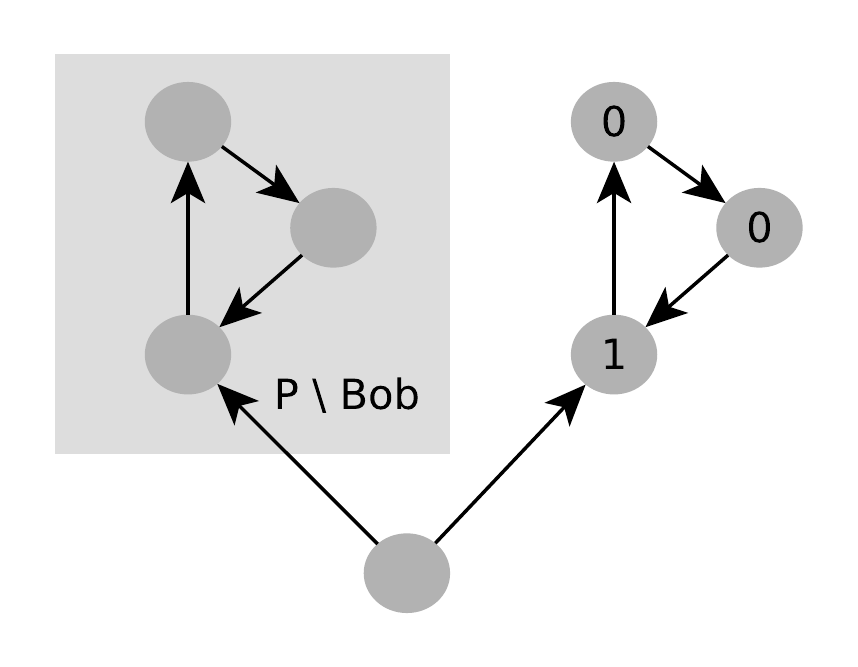}
\subcaption{Identification. The value of single RVs can be observed individually.}
\end{subfigure}
\hspace{0.2cm}
\begin{subfigure}{0.4\textwidth}
\vspace{0.7cm}
\centering
\includegraphics[scale=0.5]{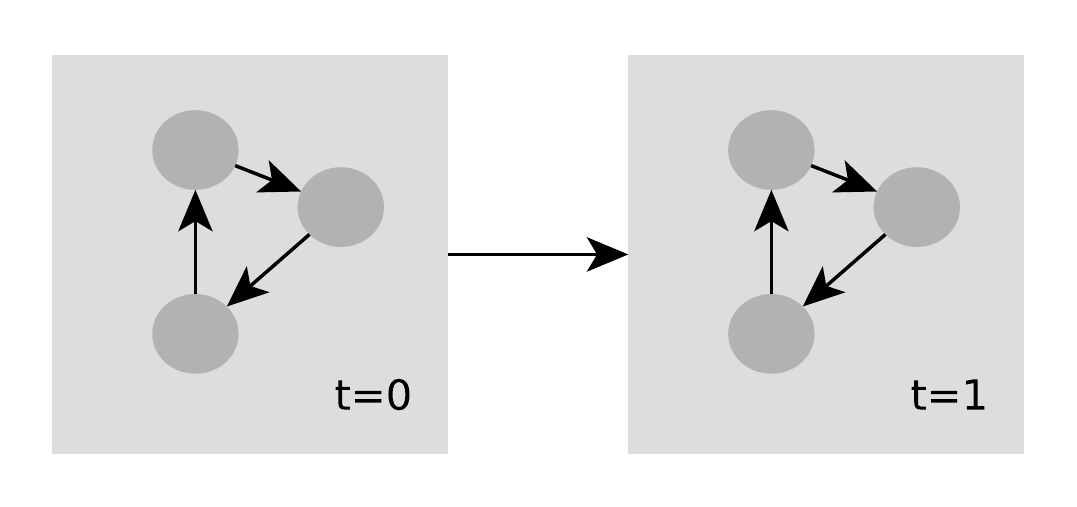}
\subcaption{Online. For each time $t$, a query is answered, each depending on current observations and estimate at time $t-1$.}
\end{subfigure}
\caption{Schematic depiction of properties of the algorithmic approaches.}
\label{fig:properties}
\end{figure}

In the following, we present six properties that characterize the algorithms we investigate in this review. 
They have been obtained by analyzing the application domains of the approaches retrieved by the systematic literature review described in Section \ref{sec:methods}.
Thus, they are a \emph{result} of the systematic review, and one of the major contributions of this review. We chose to present them at this point in the paper because they are also used as a basis for analyzing and discussing the retrieved papers. They are depicted schematically in Figure \ref{fig:properties}.

\paragraph*{Can the algorithm handle equivalent RVs efficiently as a group? (Group Variables)}
The first two properties characterize the type of abstraction that the algorithms are using. In Section \ref{sec:preliminaries}, we presented two abstraction approaches: The first one groups multiple equivalent variables and reasons over them as a group -- as for example done prominently in lifted graphical models.
For example, the RVs $c(a)$ and $c(b)$, as well as the corresponding factors $\beta_1$ and $\beta_2$  in Example \ref{ex:smokersbn} have been grouped.

\paragraph*{Can the algorithm handle distributions at the parametric level? (Parameterization)}
The second type of abstraction represents a distribution compactly
 by noting that the distribution follows some parametric form, and that it is sufficient to store and manipulate the parameters (which are typically far less than the enumeration of all values). The Kalman filter is a good example for this concept.
The parametric distributions can also make up only some factors of the joint distribution, like in the Rao-Blackwellized particle filter, or it might be necessary to consider \emph{mixtures} of parametric functions. In Example \ref{ex:smokersbn}, the $\alpha$ factor could be represented parametrically.

\paragraph*{Can the algorithm obtain a more specific distribution representation? (Splitting)}
 We identify two basic operations that can be performed by an inference algorithm to modify the degree of abstraction: \emph{Merging} and \emph{Splitting}. 
Splitting is the process of obtaining a more specific (propositional) representation from an abstract representation (in logic, this operation is known as \emph{grounding}).
Splitting operations are necessary for incorporating observations: Evidence about an RV makes this RV distinct from other RVs that are part of the same par-RV, and thus requires a \emph{split} of this par-RV and the corresponding parfactors. It can also be necessary to ensure the applicability of certain inference operations (e.g.\ the inversion elimination step in first-order variable elimination requires certain conditions that are ensured by splitting).
In Example \ref{ex:smokersbn}, if we obtain the information that Bob smokes, but we have no information whether Alice smokes, the corresponding par-RV $s(X)$ cannot be maintained any longer and has to be split into separate RVs $s(a)$ and $s(b)$.

\paragraph*{Can the algorithm obtain a more abstract distribution representation? (Merging)}
Merging (or lifting) is the reverse process to splitting: Obtaining a more abstract or aggregated representation, by grouping equivalent variables.
For example, grouping the RVs $s(a)$ and $s(b)$ into the par-RV $s(X)$ is a merging operation.
Merging is necessary in all domains where either the problem is given in a propositional form, or domains where the problem degenerates over time by repeated splitting operations. 
Splitting and merging only change the \emph{representation} of a distribution, they do not change the distribution itself (or at least, when approximate methods are used, they try to change it as little as possible).

\paragraph*{Can the algorithm handle information about individuals? (Identification)}
\new{
In lifted models, a common problem is how information about single individuals (i.e.\ single RVs) is handled. For example, suppose that in the parfactor graph given in Figure \ref{fig:smokers-parfactor}, we are provided with the evidence that Alice has cancer. In this case, the evidence can be incorporated into the model by splitting the representation, and handling Alice differently from the rest of the population. 
Not all algorithms handle identifying information by splitting. For example, when the model is given in propositional form and merging operations are applied to it, the evidence can be considered there, leaving Alice as a special case. 
Some methods do not allow to process evidence about individuals at all, like Multiset Rewriting Systems.
}

\paragraph*{Can the algorithm perform inference in dynamic domains? (Online)}
This property describes the difference between probabilistic inference and Bayesian filtering.
Probabilistic inference answers a single query (i.e. it estimates the state of hidden variables) for a \emph{single} point in time, given evidence.
Bayesian Filtering answers a \emph{sequence} of queries, one for each time step. Each query depends on the current observation, and the distribution of the hidden variables of the previous point in time.
In general, the observation sequence is not known in advance, but more observations are obtained as time passes.
 As explained in Section \ref{subsec:DBN}, such problems cannot be solved efficiently with non-sequential inference algorithms. Instead, the inference algorithms require a property that we call \emph{online inference}: Calculating the posterior probability in a sequential fashion, with a time complexity of each step that does not depend on the total sequence length. This way, observation sequences of indeterminate length can be processed by the algorithm. \\

The properties describe the application domain of the approaches:
Two approaches that are similar regarding these properties can (in principle) be applied to the same class of problems, while exploiting \emph{some} symmetry of the domain. 
We want to point out that only two of the properties describe \emph{state space abstraction} methods -- the others describe transformations between abstract and explicit representation, and further properties that are required by some domains.
They are chosen in such a way that they are meaningful for all of the approaches considered in this review\footnote{For example, Lifted Inference algorithms can be distinguished based on their algorithmic ideas (search-based, graph manipulation-based, MCMC-based etc.), the representation formalism, etc., but such a distinction is (1) not meaningful for some approaches, e.g. for Multiset Rewriting Systems, and (2) does not characterize the problem domain, as intended by us.} -- but for each resulting class of approaches, we incorporate a discussion of group-specific properties, whenever necessary.
Note that the properties do not describe complexity classes -- in contrast to the classification proposed by \shortciteA{jaeger_liftability_2012} (see Appendix \ref{sec:li-complexity-classes}), which is, however, only meaningful for a subset of all approaches, namely lifted inference algorithms.
That is, two approaches that fall in the same group can still be different regarding the subproblems for which they are \emph{tractable}.

%% file: Input/03_Methods.tex
\section{Systematic Literature Review}
\label{sec:methods}
In the following, we describe the search and evaluation methods used in this systematic review. 
As systematic reviews are not very common in computer science, this section starts by briefly introducing the systematic review methodology. 
Afterwards, we describe  how each of the steps has been realized for this review.

A systematic literature review aims at finding all relevant work addressing a specific research problem by performing a reproducible and objective process. Compared to an unstructured review, a systematic review gives a broader, unbiased view of the topic. Unstructured reviews have a higher chance to miss out contributions, either because they have not been found or because of \emph{narrative distortion}, the observations that the author of a review is more likely to include a paper if it supports the argumentation structure of the review.
A systematic review consists of the following steps \cite{kitchenham_procedures_2004}:
(1) define the research question, (2) define the search procedure, (3) identification of research items (papers), (4) paper selection, (5) paper analysis.
The PRISMA statement \shortcite{moher_preferred_2009} is an established guideline that describes which items should be reported in a systematic review. In this review, we try to follow this guideline whenever possible. However, the PRISMA statement is directed towards quantitative analysis of medical research, whereas the present review is more concerned with qualitative aspects, namely assessing solution strategies to a specific problem. Therefore, some items could not be reported. 

The research question (of the systematic review, not to be confused with the research question of the analyzed papers) typically consists of the following parts: (1) What research exist that solve problem P? (2) How are the solutions of P related to each other? (3) What further research topics arise from the existing research?
After the research question is made clear, the search procedure to answer this question is defined. This includes the definition of search terms as well as the publication databases that are used for the literature search.
A common strategy to identify search terms is to use a set of pilot papers that are known to be relevant, based on prior knowledge of the field. These pilot papers then guide the definition of the search terms, by making sure that all of them are retrieved.

Based on the search terms, the selected publication databases are searched and a list of initial papers is retrieved. These papers are then examined to assess their relevance to the research question, based on predefined \textit{inclusion} and \textit{exclusion criteria}. This step is performed by only considering the title, abstract and keywords of each paper.
Afterwards, the full-text of the remaining papers is retrieved and their relevance regarding the inclusion and exclusion criteria is examined once again. The remaining papers are called \textit{primary papers}.
The primary papers are then analyzed with respect to the research question. This includes finding the underlying structure and relationship of the approaches and identifying possible research gaps.
In the following, we describe how each of the steps has been implemented for this review.

\subsection{Research Question}
As described in the introduction, this review aims at giving an overview over solutions to the state space explosion problem from different research fields. More specifically, we are concerned with probabilistic inference algorithms that exploit state space abstractions.
 Our goal is to identify the common underlying structure of the approaches: 
 What are common properties of the algorithms, and how does this reflect their capabilities, i.e. their applicability to different problem instances.

More formally, these questions can be stated as follows:
\begin{itemize}
\item[\textbf{Q1}] What methods exist to overcome the state space explosion problem in probabilistic inference? 
\item[\textbf{Q2}] What types of problems can different methods be applied to, and how is this reflected by the properties of the methods?
\item[\textbf{Q3}] How are these methods related to each other, i.e.\ are similar concepts used in multiple approaches?
\item[\textbf{Q4}] Which topics for future research can be derived?
\end{itemize}

\subsection{Search Procedure}
For the literature search, we used the publication databases ScienceDirect, IEEE Xplore, ACM digital library, and Scopus. These databases were chosen based on their relevance for computer science publications, and the possibility to perform a search only on title, abstract and keywords of a publication\footnote{Another common publication database, SpringerLink, was not used because it only allowed full text searches.}.
 Our definition of search terms has been based on 10 pilot papers \shortcite{barbuti_maximally_2011,de_salvo_braz_lifted_2005,gogate_probabilistic_2016,huang_fourier_2009,kersting_lifted_2012,kwiatkowska_symmetry_2006,milch_lifted_2008,niepert_markov_2012,poole_first-order_2003,singla_lifted_2008} that were the result of an explorative investigation of the literature.
  The search terms were defined to make sure that all of these papers have been retrieved. 
However, they were formulated in a general way and do not aim at specific papers or methods, to retrieve as many papers as possible that are relevant for the scope of this review. 
  The search terms have been iteratively refined during the search process, by adding search terms to the set whenever we discovered literature that we considered relevant, and the field has not been fully covered by the current terms. The resulting terms are shown in Table \ref{table:searchterms}.

\begin{table}
\centering
\begin{tabular}{l}
\toprule
First term set\\
\midrule
lifted\\
first order\\
higher order\\
symmetry\\
permutation\\
multiset\\
\bottomrule
\end{tabular}
\hspace{0.5cm}
\begin{tabular}{l}
\toprule
Second term set\\
\midrule
bayesian inference\\
probabilistic inference\\
probabilistic reasoning\\
graphical model\\
bayesian network\\
state space model\\
recursive bayesian estimation\\
bayesian filtering\\
particle filter\\
hidden markov model\\
probabilistic multiset rewriting\\
multi-agent\\
multi-target\\
multi-object\\
activity recognition\\
plan recognition\\
\bottomrule
\end{tabular}
\caption{Search terms used to construct search query.}
\label{table:searchterms}
\end{table}

The first term set describes possible state space abstractions, the second term set describes the domain where the abstractions are applied, or the research area where such abstractions are used. 
We constructed the query by connecting all terms in a set with logical OR and both sets with logical AND. This query describes all papers where at least one of the terms of the first set and at least one of the elements of the second set occurs. The search has been performed on the title, keywords and abstract of the publications. 
This way, the number of results stayed manageable, and we still retrieved all papers where any of the terms occurred prominently (i.e. that might be relevant).

\subsection{Paper Selection}
The search results have been assessed based on the following inclusion criteria. 
\begin{enumerate}
\item[\textbf{I1}] The paper is written in English. 
\item[\textbf{I2}] The paper is peer-reviewed.
\item[\textbf{I3}] The full text of the paper is available via IEEExplore, the ACM Digital Library, SpringerLink, ScienceDirect, or other sources like the author's website.
\item[\textbf{I4}] The paper includes a novel algorithmic contribution.
\item[\textbf{I5}] The paper is considering a probabilistic model.
 \item[\textbf{I6}] The paper presents an inference algorithm for the probabilistic model.
 \item[\textbf{I7}] The paper presents an abstract representation of the state space or a method to reduce the state space.
 \item[\textbf{I8}] The inference algorithm exploits the state space abstraction.
\end{enumerate}
Criteria \textbf{I1}-\textbf{I3} make sure that the analysis of the papers is feasible for us.
This review focuses on technical approaches to handle the state space explosion problem. 
Therefore, \textbf{I4} ensures that application and review papers are excluded. 
Criterion \textbf{I5} implies that only approaches that model a probability distribution have been considered. 
Reduction methods in \emph{deterministic} settings, like first-order resolution, or state space abstraction in search problems \cite{holte_state_2015}, were excluded by this criterion:
Although they might contain interesting ideas on how a state space can be abstractly represented, they cannot be applied to probabilistic models in a straightforward manner. 
For \textbf{I6}, we defined probabilistic inference as \emph{calculating a posterior distribution, given a prior distribution}. This definition also includes inference algorithms for dynamic domains, that might perform this step repeatedly.
Criteria \textbf{I7} and \textbf{I8} ensure that only approaches that exploit a state space reduction method were included. Specifically, approaches that perform inference by grounding the abstract representation were not included, for example approaches known as knowledge-based model construction.
The rationale is that this review is focused on inference algorithms that actually exploit the lifted representation, i.e. directly reason in the lifted domain.

Paper inclusion/exclusion used a three-step process. At first, only the title, abstract, and keywords of each publication have been examined.
The full-text of the remaining papers has been examined in more detail. 
By examining the references in the remaining papers, we identified additional relevant papers (see flow diagram in Figure \ref{fig:paper-dataflow}).

\subsection{Analysis Procedure}
\label{subsec:analysisprocedure}

We analyzed the remaining papers in order to answer research questions \textbf{Q1} -- \textbf{Q4}.
The analysis is based on the \emph{properties of inference algorithms} defined in Section \ref{sec:properties}, i.e.\ these properties have been assessed for all approaches described in the retrieved papers.
Afterwards, we performed a clustering of the approaches based on their manifestation of the properties, i.e.\ all approaches having the same manifestation of the properties form a cluster (or group). 
These groups thus define all approaches that behave similar from an application point of view, i.e.\ all approaches from the same group can be applied in the same problem domain (although different subclasses of the domain may be solvable efficiently).

%% file: Input/04_Results.tex
\begin{figure}[tb]
\centering
\includegraphics[scale=0.5]{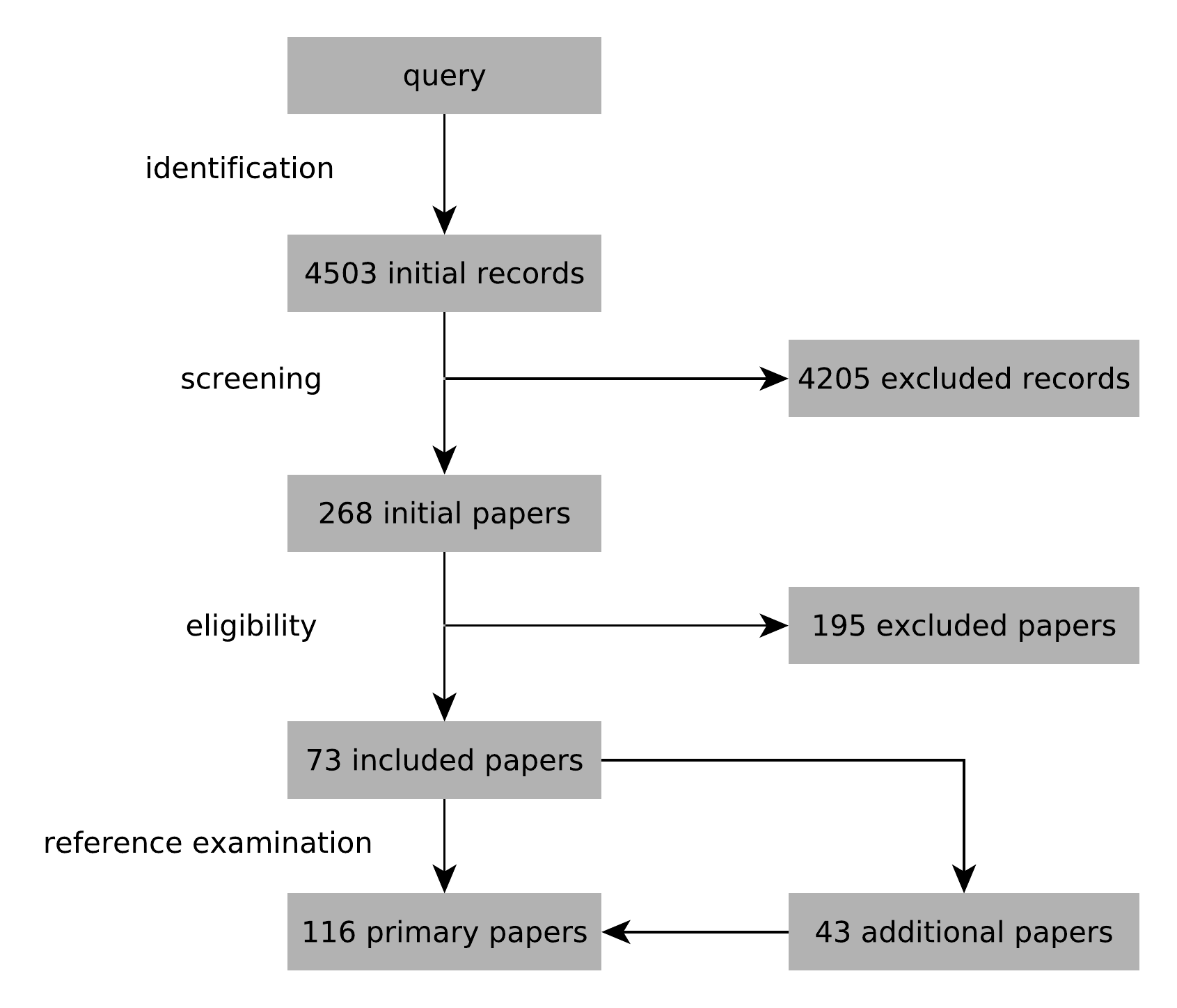}
\caption{Flow Diagram of paper selection, oriented on PRISMA statement \shortcite{moher_preferred_2009}.}
\label{fig:paper-dataflow}
\end{figure}

\begin{figure}[tb]
\centering
\includegraphics[scale=0.75]{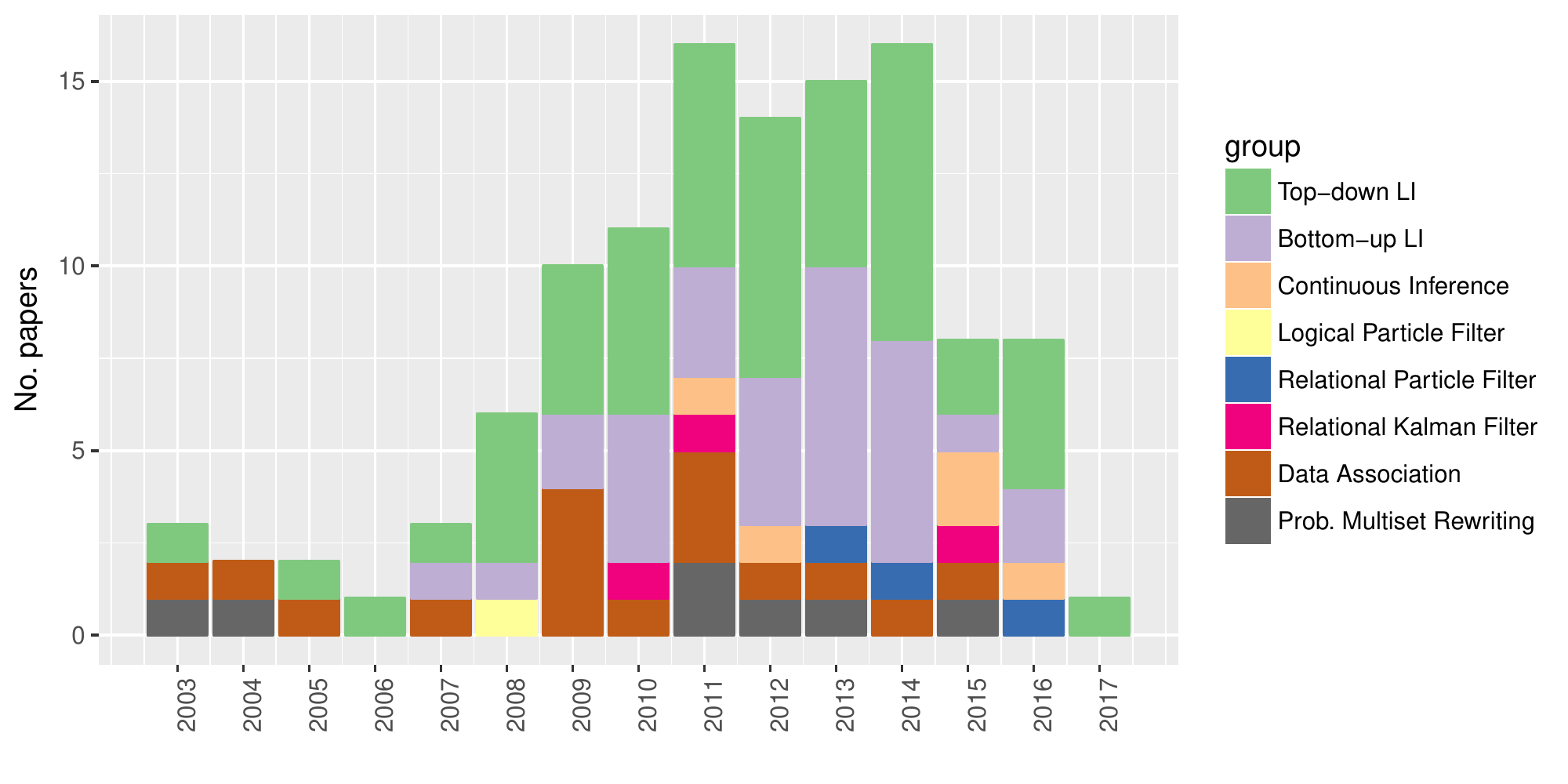}
\caption{Number of examined papers per year.
The papers have been retrieved from January to February 2017. The groups are based on the analysis and clustering of approaches, as described in the text.}
\label{fig:literature-hist}
\end{figure}

\subsection{Results}
\label{sec:results}

\begin{table}[tb]
\begin{tabular}{llp{14cm}}
\toprule
Crit. & \# & Explanation \\
\midrule
\textbf{I1} & 3 & Paper not written in English \\
\textbf{I2} & 0 & Full-text not available\\
\textbf{I3} & 9 & Paper not peer reviewed \\
\textbf{I4} & 31 & Paper does not contain a novel algorithmic contribution (e.g. application and review papers) \\ 
\textbf{I5} & 11  & Model is not probabilistic (e.g. inference in first-order logic) \\
\textbf{I6} & 77 & No inference algorithm for probabilistic model (e.g. because paper presents an algorithm for learning the model structure, or something completely different, like planning or model checking) \\
\textbf{I7} & 17 & No lifted representation of probabilistic model (e.g. propositional models) \\
\textbf{I8} & 46 & Inference algorithm does not exploit abstract representation (e.g. it relies on a complete grounding) \\
\bottomrule
\end{tabular}
\caption{Reasons for excluding 195 of the 268 papers that remained after examining title, keywords and abstract of the 4503 initial records.}
\label{tbl:reasons-for-exclusion}
\end{table}

This section gives quantitative results about the retrieved papers.
From the 4503 initial records that have been retrieved by the database search, 4235 have been excluded by only examining their title, keywords, and abstract.
The relevance of the remaining 268 papers (regarding the inclusion criteria) has been examined based on the full-text. 195 of those papers have been excluded, based on the inclusion criteria as shown in Table \ref{tbl:reasons-for-exclusion}.
When multiple reasons apply to one paper, it is grouped under the the first reason, based on the order of the inclusion criteria.
The high number of papers excluded because of \textbf{I6} shows that the query terms have been chosen very broadly, such that also a great number of papers that are not concerned with probabilistic inference have been retrieved. Most of the papers excluded because of \textbf{I8} are concerned with \emph{knowledge-based model construction}, i.e.\ propositional inference in lifted models, a research field much older than lifted inference.
In Appendix \ref{subsec:related}, it is further discussed why specific approaches that might seem relevant have not been included.

The remaining 73 papers were considered relevant and included into this review. The references of these papers were examined, which lead to the identification of another 43 relevant papers.
Thus, 116 papers have been included in this review in total. This corresponds to a precision of $73/4503=1.6 \%$ and a recall of $73/116=62.9 \%$ of the initial query.
These low values point to the fact that the terminology in the field is not very consistent.

The properties of the approaches presented in these 116 papers have been evaluated, as described in Section \ref{subsec:analysisprocedure} (thus answering \textbf{Q1}).
We then clustered the approaches, as described in Section \ref{subsec:analysisprocedure}: All approaches having the same manifestation of the properties have been put into the same group. With this process, we found eight distinct groups. We assigned names to the groups that seemed appropriate to us. The groups are shown in Table \ref{table:clustering}. The complete list of all papers per group is shown in Appendix \ref{sec:paperstogroups}. 
 We want to emphasize that the groups have not been predefined, but they are a result of the individual analysis of each paper.

As can be seen from Table \ref{table:clustering}, the ``lifted inference'' groups contain by far the most papers. 
This shows that lifted inference is a very active research area. 
The other groups contain fewer papers. One reason may be that they belong to a larger research area (for example, there are numerous papers on data association in general), but only a small subset of the approaches employ state space abstraction.

Figure \ref{fig:literature-hist} shows the chronological development of the research area. 
Although the first lifted inference paper was published in 2003, the majority of lifted inference papers has been published after 2008. The drop in the total number of included papers after 2014 may be due to the fact that not all papers form 2015 and 2016 are properly indexed at the used publication databases at the time of retrieving the papers (January -- February 2017).

%% file: Input/05_Analysis.tex
\section{Categories of Inference Algorithms}
\label{sec:analysis}
\newcommand*{\greysquare}{\textcolor{gray}{\blacksquare}}
\begin{table}[t]
\centering
\input{figures/clustering-table.tex}
\caption{Groups of inference approaches, based on the properties defined in Section \ref{sec:properties}. $\blacksquare$: has property, $\square$: does not have property, -: property not necessary/not meaningful.}
\label{table:clustering}
\end{table}

As discussed in Section \ref{sec:results}, we defined groups or classes of approaches that consist of all approaches that are similar regarding the six properties defined in Section \ref{subsec:analysisprocedure} (shown in Tables \ref{table:clustering} and Appendix \ref{sec:paperstogroups}).
In the following, we briefly describe the common algorithmic ideas that are shared by all approaches in the same group.

\subsection{Lifted Inference}
\label{subsec:lifted-inference}
Lifted inference algorithms are concerned with probabilistic inference in lifted graphical models (Section \ref{subsec:liftedgraphicalmodels}). 
They aim at performing the inference directly in the first-order domain, without grounding the lifted graphical model, whenever possible. 
\new{
By maintaining the lifted representation, they can exploit the symmetries and redundancies that are inherent to these representations. More specifically, lifted inference algorithm can be seen as exploiting \emph{exchangeability} in the model \cite{niepert_tractability_2014}: They exploit the fact that in lifted graphical models, it is not necessary to know the \emph{specific} RVs having a certain value, but only the \emph{number} of RVs having each value. In general, lifted inference algorithms can be viewed as performing the following steps: (1) Decompose the inference problem into similar, independent subproblems, (2) solve one representative instance, (3) count the number of instances (instead of generating all instances) \cite{taghipour_completeness_2013}.
}

How these steps are implemented is specific to the different lifted inference algorithms. As a high-level distinction, we distinguish between top-down and bottom-up lifted inference, following \shortciteA{kersting_lifted_2012}. 
The difference of these approaches is the input they receive: Top-down lifted inference algorithms start with a lifted graphical model, while bottom-up algorithms receive a propositional model as input \new{(thus, they are different in step (1) -- the generation of subproblems)}. 
\new{From the algorithmic viewpoint, this distinction is not always very precise, as it is just a matter of preprocessing: For several algorithms, both top-down and bottom-up versions exist -- for example, lifted belief propagation has top-down \shortcite{singla_lifted_2008} and bottom-up \cite{kersting_counting_2009} variants.
However, as this review is explicitly concerned with the \emph{problem class} each approach can process, we still consider bottom-up/top-down a meaningful distinction -- it is also directly reflected by the \emph{properties} (Section \ref{sec:properties}) of the algorithms:}
Top-down algorithms apply \textsf{splitting} operations, while bottom-up algorithms need to perform \textsf{merging} operations on the propositional model (but never need to apply splitting operations)\footnote{\new{Search-based algorithms are also considered top-down. They branch on the value of the (par-)RVs, resulting in a simpler inference problem in each branch. We consider this branching a form of \textsf{splitting}.}}. 
\new{Top-down algorithms, on the other hand, never apply merging operations (i.e.\ they never explicitly search exchangeable RVs and group them)}.

We want to point out that lifted inference problems and algorithms can be structured further, as proposed by \shortciteA{jaeger_liftability_2012}: Broadly speaking, the idea is to classify lifted probabilistic models by  the ``complexity'' of their structure (in terms of numbers of parameters of par-RVs and parfactors). For some of the resulting classes, it can be shown that inference can always be performed in time that is polynomial in the domain size of the model, while in general, no such guarantees can be given. 
In Appendix \ref{sec:li-complexity-classes}, we give an overview over these problem classes. However, as an in-depth discussion of lifted inference is not the focus of this review, we do not discuss this classification in detail here. 
From the high-level point of view of this review, all lifted inference algorithms are concerned with a similar problem: Efficient inference in graphical models containing symmetries.
 For a more in-depth discussion, we refer to the review papers of \shortciteA{kersting_lifted_2012} and \shortciteA{kimmig_lifted_2015}, as well as the books by \shortciteA{de_raedt_statistical_2016} and \shortciteA{getoor2007introduction}.
In the following, we explain the general idea of some prominent lifted inference algorithms (first-order variable elimination, lifted recursive conditioning, lifted belief propagation).

\subsubsection{Top-Down Lifted Inference}
\label{subsubsec:top-down-li}
\paragraph*{First-order Variable Elimination}
\shortciteA{poole_first-order_2003} proposed the first ideas related to lifted inference, in an algorithm known as first-order variable elimination (FOVE).
The idea is to perform variable elimination directly on a parfactor graph, eliminating entire par-RVs in one step, instead of single RVs.
\begin{example}
Consider the graphical model of Example \ref{ex:smokersbn} and the query $P\bigs{(}s(X),d{=}1\bigs{)}$. 
Remember that inference in the propositional model (with $X=\{a,b\}$) requires two elimination steps, the elimination of $c(a)$ and $c(b)$ (Example \ref{ex:variable-elimination}). 
In the parfactor graph (Figure \ref{fig:smokers-parfactor}), we can \emph{in principle} directly eliminate the par-RV $c(X)$ by multiplying the parfactors $\beta$ and $\alpha$ and marginalizing $c(X)$ to get a factor
\begin{equation*}
f(s(X),d)=\sum_v \alpha\bigs{(}c(X){=}v,d\bigs{)}\ \beta\bigs{(}s(X),c(X){=}v\bigs{)}
\end{equation*}
which can be represented by the table
\begin{table}[H] 
\centering
\begin{tabular}{lll}
\toprule
s(X) & d & $f$\\
\midrule
0  & 0 & $\beta^{(00)}\, \alpha^{(00)} + \beta^{(01)}\, \alpha^{(10)}$ \\
0  & 1 & $\beta^{(00)}\, \alpha^{(01)} + \beta^{(01)}\, \alpha^{(11)}$ \\
\vdots  & \vdots & \vdots \\
\bottomrule
\end{tabular}
\end{table}
This factor directly leads to the query solution $P\bigs{(}s(X),d{=}1\bigs{)}=f\bigs{(}s(X),d{=}1\bigs{)}$. 
\end{example}
The elimination step performed for eliminating $c(X)$ in the example is called \emph{inversion elimination}.
\new{Not all cases can be handled this way: For example, consider the case of eliminating $d$: In the ground factor graph, eliminating $d$ means we need to multiply all $\alpha_i$ factors, resulting in a factor of all $c(X)$, i.e.\ a factor that has exponential size with respect to the domain. In general, inversion elimination can only be applied when the parameters that appear in the par-RV to be eliminated are the same as the parameters in each parfactor depending on this par-RV. Thus, for eliminating $d$, inversion elimination cannot be applied, and FOVE as proposed by \shortciteA{poole_first-order_2003} needs to ground $c(X)$ and create the the exponentially large factor. 
}

\new{
However, the RV $d$ (whether or not a death occurred last year) might only depend on the \emph{number} of people having cancer, not their specific identities. Thus, it is sufficient that the resulting factor considers the number of instances of $c(X)$ that are true. This was first realized by \citeA{de_salvo_braz_lifted_2005}, who presented an elimination operator that can handle this case. Later, \citeA{milch_lifted_2008} proposed an explicit representation of such factors, called \emph{counting formulae}, that have later been generalized by  \shortciteA{taghipour_generalized_2014}.
Additional elimination rules that make FOVE applicable to more cases without grounding are provided by \shortciteA{apsel_extended_2011}, and \shortciteA{taghipour_generalized_2014,taghipour_completeness_2013-1}. 
Using these rules, the class $FO^2$ (inference problems containing at most two parameters per parfactor, see Appendix \ref{sec:li-complexity-classes} for more details) can always be solved in polynomial time in the parameter domain size.
The works of \shortciteA{taghipour_lifted_2012,taghipour_lifted_2013} allow for more general constraints in the parfactors.
}

\paragraph*{Lifted Recursive Conditioning}
\new{
Approaches based on variable elimination have the problem that they need to represent the intermediate results of the elimination operations, that can become increasingly complex during inference. Recently, \emph{search-based} lifted inference algorithms have emerged, that do not manipulate the representation of the parfactors directly, but branch on the values of par-RVs and combine the results of each branch appropriately.
The convenient property of these algorithms is that the intermediate results (partially instantiated lifted graphical models) become simpler with each operation, instead of more complex.
}

\new{
For example, lifted recursive conditioning \cite{poole_towards_2011} works similar to recursive conditioning (see Section \ref{subsubsec:inference-algs}), but branches on the values of par-RVs instead of (propositional) RVs. 
The algorithm exploits a similar idea as counting elimination: There are cases where it is sufficient to branch on the \emph{number} of RVs having each possible value, instead of all assignments of the RVs.
An extension of lifted recursive conditioning \cite{kazemi_new_2016} is able to solve all problems in the class $S^2RU$ in polynomial time  -- which is currently one of the largest classes where tractable inference can be guaranteed (see Appendix \ref{sec:li-complexity-classes} for details).
} 

\begin{example}
\new{
Consider the graphical model of Example \ref{ex:smokersbn} and the query $P\bigs{(}s(X),d{=}1\bigs{)}$.
At the beginning, only $d{=}1$ is instantiated and the algorithm needs to branch. Instead of branching on the values of a single RV, it creates one branch for each \emph{histogram} of possible values of the instances of a par-RV. In this example, the algorithm chooses $c(X)$ to branch, leading to three recursive calls of the algorithm, where 0, 1 or 2 instances of $c(X)$ are true, respectively. 
In each branch, the factor $\alpha$ can be evaluated. For example, in the branch where 2 instances of $c(X)$ are true, it is evaluated as $\alpha\bigs{(}d{=}1,c(X){=}1\bigs{)}^2$.
Afterwards, a similar branch is performed on $s(X)$.
Note that the result of each branch needs to be multiplied by $\binom{n}{i}$ (where $i$ is the number of true instance in the branch, and $n$ is the population size), as this is the number of equivalent ground assignments represented by this branch.
 Compared to recursive conditioning, where we need to branch on each (ground) RV, fewer branches need to be performed. 
}
\end{example}

\new{
Several other search-based algorithms have been devised. The approaches of 
Van Den Broeck et al. (2011),
 and \shortciteA{gogate_probabilistic_2016} transform the problem into a weighted model counting problem on a first-order logical theory (WFOMC): 
Given a first-order logical theory $T$ and positive and negative weight functions $w$ and $\overline{w}$ for each predicate. 
WFOMC is the problem of computing
\begin{equation}
\sum_{I \models T} \prod_{a\in I} w(\text{Pred}(a)) \prod_{a\in \text{HB}(T) \setminus I} \overline{w}(\text{Pred}(a))
\end{equation}
where $I$ is a model of $T$, $\text{HB}(T)$ is the Herbrand base of $T$ and $\text{Pred}$ is the predicate of an atom $a$.
  Note that the weighted theory defined above is different form an MLN, where a weight is assigned to each formula, not to each predicate. 
 Given a parfactor graph, one can construct a weighted theory such that the weighted model count is the probability of some evidence in the factor graph. The basic idea is that each model relates to a value assignment of the RVs that is consistent with the evidence. 
WFOMC can be solved directly by a search-based algorithm \shortcite{gogate_probabilistic_2016}, or by compilation into a first-order arithmetic circuit \shortcite{van_den_broeck_lifted_2011}.

\shortciteA{jha_lifted_2010} propose a rewriting-rule based inference algorithm. These rules take an MLN and express it as a combination of multiple simpler MLNs until the MLNs are trivial such that the solution can be computed directly.
}

\paragraph*{Probabilistic Databases}
\new{
Ideas related to lifted inference arose independently in the probabilistic database community. Probabilistic databases are relational databases where each tuple is a boolean random variable, and database queries output a probability distribution of possible answers (instead of a single answer, as in conventional relational databases). Thus, query evaluation in a probabilistic database is a probabilistic inference task. More details on probabilistic databases and query evaluation are provided in the book by \citeA{suciu_probabilistic_2011}.
}

Answering queries in probabilistic databases corresponds to an \emph{asymmetric} weighted model counting task, where weights of predicates can vary, depending on the domain constant (as compared to symmetrical WFOMC defined above, where each predicate always has the same weight). Still, symmetries can be present, allowing to use methods closely related to (bottom-up) lifted inference algorithms  \cite{sen_exploiting_2008}.
\citeA{dalvi_efficient_2007} present an algorithm that rewrites a probabilistic database query in terms of combinations of simpler queries, until trivial queries can be answered directly. This approach is conceptually similar to search-based lifted inference algorithms like lifted recursive conditioning \shortcite{poole_towards_2011}.
Typically, probabilistic databases assume that tuples are independent, which can make inference much easier in certain cases. \citeA{jha_probabilistic_2012} show how correlations can be modeled in tuple-independent databases, allowing to use lifted methods devised for tuple-independent databases \shortcite<e.g.,>{dalvi_efficient_2007} in a more general setting. 
\citeA{dylla_top-k_2013} devise an algorithm for finding the $k$ most probable query results according to their marginal probabilities, without the need to first materialize all answer candidates.
The symmetrical case of WFOMC has also been considered by the probabilistic database community, leading to new insights on domain-liftable inference problem classes  \cite{beame_symmetric_2015} -- outlined in Appendix \ref{sec:li-complexity-classes}.

\subsubsection{Bottom-Up Lifted Inference}

\begin{figure}[t]
\centering
\begin{subfigure}{0.18\textwidth}
\centering
\includegraphics[scale=0.55]{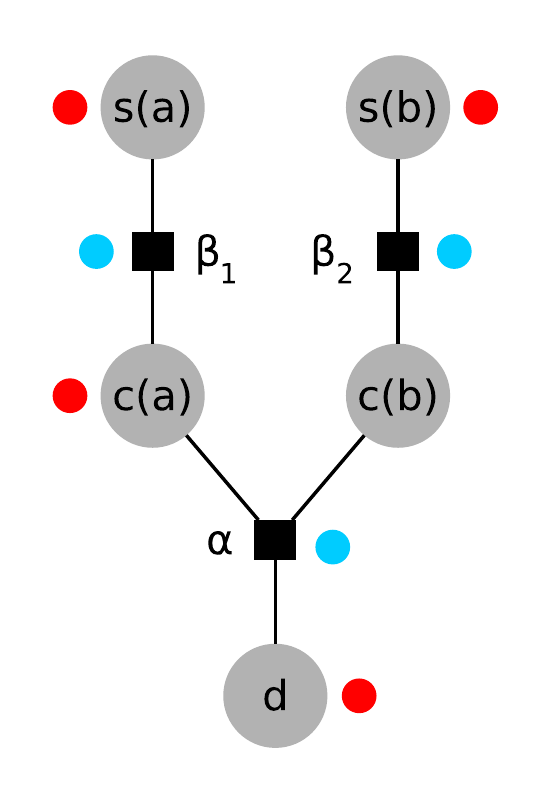}
\captionsetup{justification=raggedright,
singlelinecheck=false
}
\caption{Initially, all RV and factor nodes have same color.}
\end{subfigure}
\hspace{0.01\textwidth}
\begin{subfigure}{0.18\textwidth}
\centering
\includegraphics[scale=0.55]{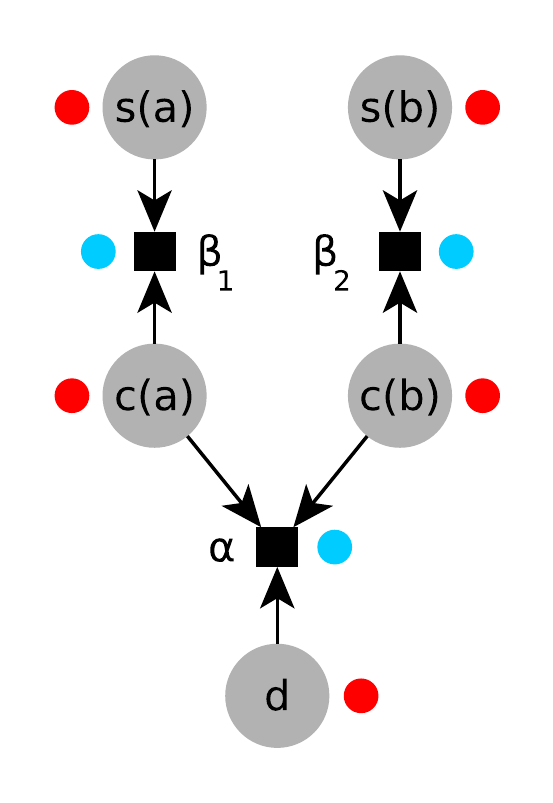}
\captionsetup{justification=raggedright,
singlelinecheck=false
}
\caption{RV nodes send color to factor nodes.}
\end{subfigure}
\hspace{0.01\textwidth}
\begin{subfigure}{0.18\textwidth}
\centering
\includegraphics[scale=0.55]{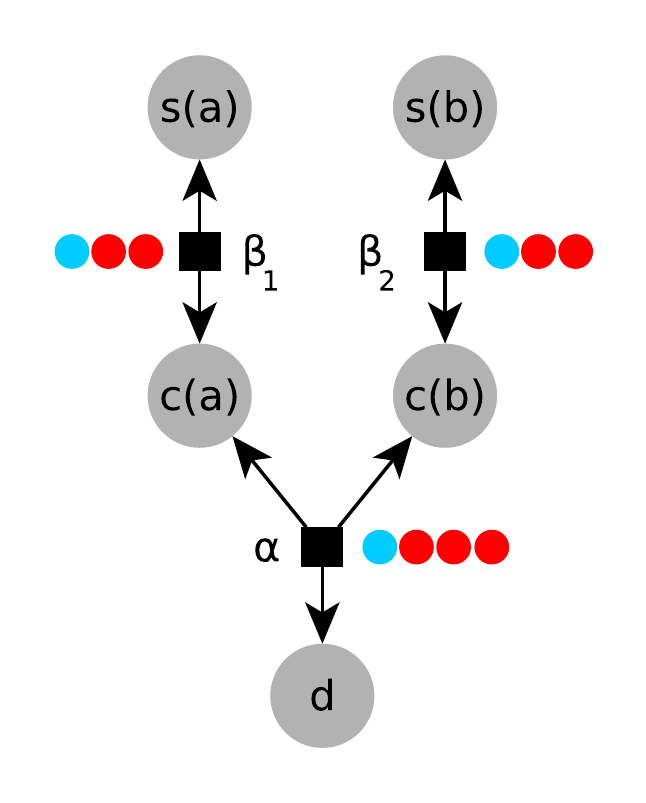}
\captionsetup{justification=raggedright,
singlelinecheck=false
}
\caption{Factor nodes send stacked colors to RV nodes.}
\end{subfigure}
\hspace{0.01\textwidth}
\begin{subfigure}{0.18\textwidth}
\centering
\includegraphics[scale=0.55]{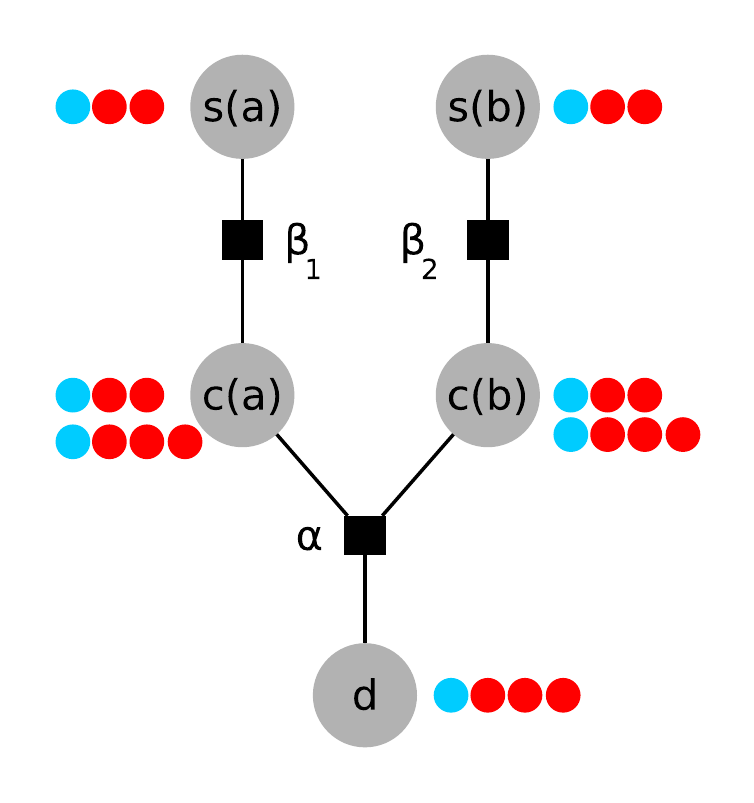}
\captionsetup{justification=raggedright,
singlelinecheck=false
}
\caption{Colors are stacked at RV nodes.}
\end{subfigure}
\hspace{0.01\textwidth}
\begin{subfigure}{0.17\textwidth}
\centering
\includegraphics[scale=0.55]{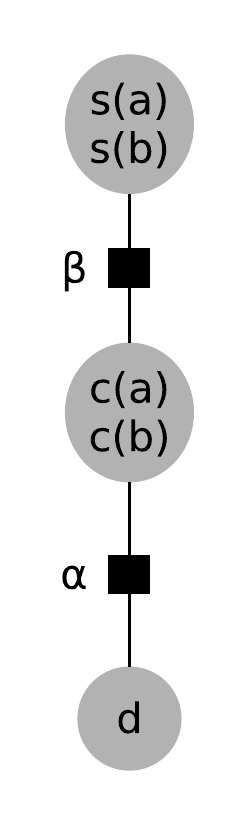}
\captionsetup{justification=raggedright,
singlelinecheck=false
}
\caption{RV nodes with same colors are grouped.}
\end{subfigure}
\caption{From left to right, the steps of lifted BP factor graph compression \shortcite<adapted from>{kersting_counting_2009}.}
\label{fig:lbp-example}
\end{figure}

As opposed to top-down lifted inference algorithms, bottom-up approaches take a propositional model and perform \textsf{merging} operations to obtain a first-order structure that can be exploited.
Thus, bottom-up approaches are potentially applicable to a larger class of problems, as they do not require the model to be in lifted form (or to even contain exact symmetries, instead they can approximate the model by a symmetric one).
 However, performing \textsf{merging} operations is an additional overhead: The propositional model can be very large, and merging requires at least linear time in the propositional model size.

A well-known bottom-up lifted inference algorithm is lifted belief propagation proposed by \shortciteA{kersting_counting_2009}. 
The idea is to perform belief propagation (BP) on a factor graph where each node represents a set of nodes that would send and receive the same messages in standard BP. 
This lifted factor graph is obtained by simulating BP and keeping track of which nodes send and receive the same messages.
In this simulation, each node sends its color (a signature) instead of the actual message. Initially, all RV and factor nodes have the same color signature. The colors a node receives extend the current color of the node. This color signature is sent in consecutive messages. After one iteration (all nodes have sent and received a message), nodes with the same color signature are grouped for the next iteration. 

\begin{example}
\label{ex:bottom-up-li}
Figure \ref{fig:lbp-example} shows the steps of simulating BP and compressing the factor graph of Example \ref{ex:smokersbn}. The nodes $s(a)$ and $s(b)$, $c(a)$ and $c(b)$ as well as $\beta_1$ and $\beta_2$ have the same color signature after one iteration. Thus, they are grouped together in the factor graph.
Afterwards, a modified BP algorithm is performed on the compressed factor graph. This algorithm needs to consider the actual number of messages sent and received by the grouped nodes. For example, a message sent from node $c(a),c(b)$ to $\alpha$ actually represents two identical messages.
\end{example}

\new{
For cases where it is necessary to answer multiple queries on the same graphical model with only slight changes in the evidence, it is not necessary to re-construct the lifted network from scratch each time. Instead, \citeA{nath_efficient_2010}, and \citeA{ahmadi_lifted_2010} showed how the lifted network can be re-used, which is not trivial, as the structure of the lifted network depends on the evidence.
These methods can be used to realize lifted variants of the Kalman filter and PageRank algorithm \cite{ahmadi_lifted_2011}, as well as lifted linear program solvers \cite{mladenov_lifted_2012}.
}

Other bottom-up algorithms find symmetries in the graphical model by examining graph automorphisms of the graphical model. These automorphisms can be used for lifted variational inference \cite{bui_automorphism_2013} and lifted sampling-based inference \cite{niepert_markov_2012,venugopal_lifting_2012}.
In general, such approximate algorithms (based on sampling or belief propagation) can be feasible for very large and complex models, where exact inference (like variable elimination or recursive conditioning) is infeasible. 

Another interesting property of bottom-up algorithms is that they can potentially also be applied to models that are not exactly symmetric, but exhibit \emph{approximate} symmetries. This can happen when evidence about individuals is observed, and is a main issue for exact lifted inference algorithms. Methods have been devised that that approximate the model by a symmetric one, and then perform lifted inference in the symmetric model \cite{singla_approximate_2014,venugopal_evidence-based_2014,van_den_broeck_lifted_2015}. Combining this with approximate inference algorithms can lead to even more efficient inference.

\subsection{Continuous Inference}
\label{subsec:cont-inference}
\new{
Most research on probabilistic inference is concerned with discrete RVs, although many practical problems require modeling continuous variables. 
For inference in graphical models containing continuous RVs, algorithms for discrete models cannot be used directly, as they typically rely on enumerating all values of the RV. 
Instead, it is necessary to describe the functional form of the factors containing continuous RVs and manipulate them analytically (this is an instance of the \textsf{parameterization} property introduced in Section \ref{sec:properties}). Typical operations that need to be handled are marginalization (integration) and multiplication of such continuous factors.
In general, such operations can be difficult or impossible. 
However, recent research has focused on \emph{piecewise polynomial} functions for describing factors, which can be manipulated efficiently.
For example, in the approach by \citeA{sanner_symbolic_2012-1}, factors are represented as piecewise polynomial functions that are noted as case statements, as illustrated by the following example.
}

\new{
\begin{example}
\label{ex:continuous-inference}
The position of an object is observed by a noisy sensor observation. Both the position ($x$) and the observation ($o$) are continuous RVs.
The sensor can either fail, or work properly (modeled as a binary RV $b$).
In the former case, the observation density is uniform in the interval $[0,10]$ (i.e.\ the density is constant $1/10$ in this interval, such that it integrates to one).
 In the latter case, the conditional observation density is a quadratic function, centered at the real position and truncated at a distance of one from the true position\footnote{The added constant $5/6$ ensures that the density is always positive and integrates to one.}.
 This continuous distribution can be represented by a case statement as follows:
\begin{equation*}
p(o|x,b) = \begin{cases}
-(o-x)^2+5/6 &b=0 \land x-1 \leq o \leq x+1\\
1/10 &b=1 \land 0 \leq o \leq 10\\
0 &\text{otherwise}
\end{cases}
\end{equation*}
\end{example}
}

In the approach by \shortciteA{sanner_symbolic_2012-1}, inference is defined in terms of variable elimination. When a variable is marginalized from a factor (a piecewise polynomial function), the factor is integrated on the variable to be eliminated. This integration can be calculated symbolically. The resulting factor can be more complex than the original factor (i.e.\ it can contain more cases), but it is always again a piecewise polynomial function and thus can be represented by case statements. 
These operation thus result in a more complex, explicit representation of the distribution (more cases need to be distinguished explicitly) -- in the context of this review, this is a \textsf{splitting} operation.

\new{
We can also think of an operation similar to \textsf{merging} for continuous inference methods: Given a distribution as case statement, a merging operation finds an equivalent case statement with fewer cases. For example, consider the case statement
\begin{equation*}
p(a) = \begin{cases}
-a & -1 \leq a \leq 0\\
a & 0 < a \leq 1 \\
0 & \text{otherwise}
\end{cases}
\end{equation*}
where the first two cases can be merged into the single case $|a|\text{, when } -1 \leq a \leq 1$.
Such operations are implicitly performed in the approach by \shortciteA{sanner_symbolic_2012-1}, who represent case statements as some variant of algebraic decision diagrams (ADDs) -- this way, it is ensured that the case statements can be represented sufficiently compact.
}

\new{
Inference algorithms in continuous or hybrid models that rely on polynomial approximations have also been devised in the context of belief propagation \cite{shenoy_inference_2011}, and weighted model counting \cite{belle_probabilistic_2015}.
}

\subsection{Probabilistic Multiset Rewriting Systems}
Multiset rewriting systems (MRSs) \cite{calude_multiset_2001} are a formalism to model dynamic systems where the state can be described as a \emph{multiset} of entities  (i.e.\ they perform \textsf{online} inference).
 The state transitions are defined in terms of rewriting rules having preconditions (a multiset of entities that are consumed by the reaction) and effects (a multiset of entities that are created by the reaction). 
They are for instance used to model biochemical reactions \shortcite{barbuti_maximally_2011}, population dynamics in ecological studies \cite{pescini_dynamical_2006} or network protocols \cite{cervesato_meta-notation_1999}.

\newcommand\multiset[1]{\ensuremath{\llbracket\, #1 \,\rrbracket}}
\begin{example}
\label{ex:mrs-1}
A system consists of prey ($x$) and predators ($y$). Prey can reproduce, and predators can eat prey. In this simple model, eating a prey results in the death of the prey and the birth of a predator.
This system can be modeled as a MRS with the two rewriting rules $r(x) \rightarrow 2x$ and $e(x,y) \rightarrow 2y$.
\end{example}
Stochastic MRSs \cite{bistarelli_representing_2003-1} assign weights to each rule, thereby specifying the probability of selecting this rule. 
\new{
Typically, MRSs are used for simulation studies: At each step, one of the rules is sampled according to their probabilities, leading to a sequence of multiset states. 
\begin{example}
\label{ex:mrs-2}
Consider the multiset state\footnote{We use $\multiset{\cdot}$ to denote multisets.} consisting of two predators and two preys $s=\multiset{2x, 2y}$ and the rules $r(x) \rightarrow 2x$ and $e(x,y) \rightarrow 2y$ given in Example \ref{ex:mrs-1}. The rules have the weight $w_r=2$ and $w_e=1$. Thus, their probability is $p(r)=2/3$ and $p(e)=1/3$ and the successor states $s_r=\multiset{3x, 2y}$ and $s_e=\multiset{1x, 3y}$ have the same probabilities.
\end{example}
}

A popular formalism relying on MRS semantics are P Systems \cite{paun_membrane_2012},  where states can have a hierarchical structure (i.e.\ multisets can contain other multisets, and rewriting rules can also apply to the components of these inner multisets).
Instead of executing one action per time step, they define the state transitions by \emph{parallel} rule applications: At each step, a \emph{maximal} multiset of rules (i.e.\ such that no more rules are applicable at the same time step, given the multiset state) is executed.

\new{
\begin{example}
Consider the same situation as in Example \ref{ex:mrs-2}, but a parallel transition semantics.
The following maximal rules are applicable: $c_1=\multiset{2 r(x)}$, $c_2=\multiset{1 r(x), 1 e(x,y)}$ and $c_3=\multiset{2 e(x,y)}$.
 To compute the weight of each parallel rule, we multiply the weights of the individual rules and the number of ways that entities in the state can be assigned to the preconditions of the actions. 
Thus, the weights of the parallel actions are  $w_1=w_r^2 * 2 = 8$, $w_2 = w_r * w_e * 2 * 2 = 8 $ and $w_3=w_e^2 * 2 = 2$.
Finally, the probabilities are obtained by normalizing the weights: $p(c_1)=p(c_2)=4/9$, $p(c_3)=1/9$.
\end{example}
}

\new{
Computing the distribution of maximally parallel rules is a search problem related to weighted model counting (WMC): Each maximally parallel rule is a model of an appropriately defined formula. Instead of the sum of all weights of all models (as in WMC), the goal is to enumerate all models and their weights.
}

The state space representation of MRSs \textsf{groups equivalent variables}, and reasons about them as a group. 
When computing the applicable rules (and their probabilities), we only need to reason about the \emph{number} of entities of a species in a multiset, not their specific identities or ordering. This concept is related to counting formulae in C-FOVE, where probabilities only depend on the \emph{number} of RVs of a parfactor with a specific value, and not the specific identities.
For example, in the predator-prey scenario above, the probability of applying the reproduction rule depends only on the number of prey, and the probability of applying the eating rule depends only on the number of predator-prey pairs. However, the probability does not depend on presence of any specific predator or prey entity.

However, there is no way for existing MRS algorithms to reason about individual entities: 
 All entities belonging to the same species are exactly identical. From our point of view, a MRS always operates on an abstract representation, and never propositionalizes the state space (by \textsf{identification} of specific entities). 
Therefore, \textsf{splitting} and \textsf{merging} operations are not meaningful for this representation.

\subsection{Logical Particle Filter}
\begin{figure}
\centering
\includegraphics[scale=0.5]{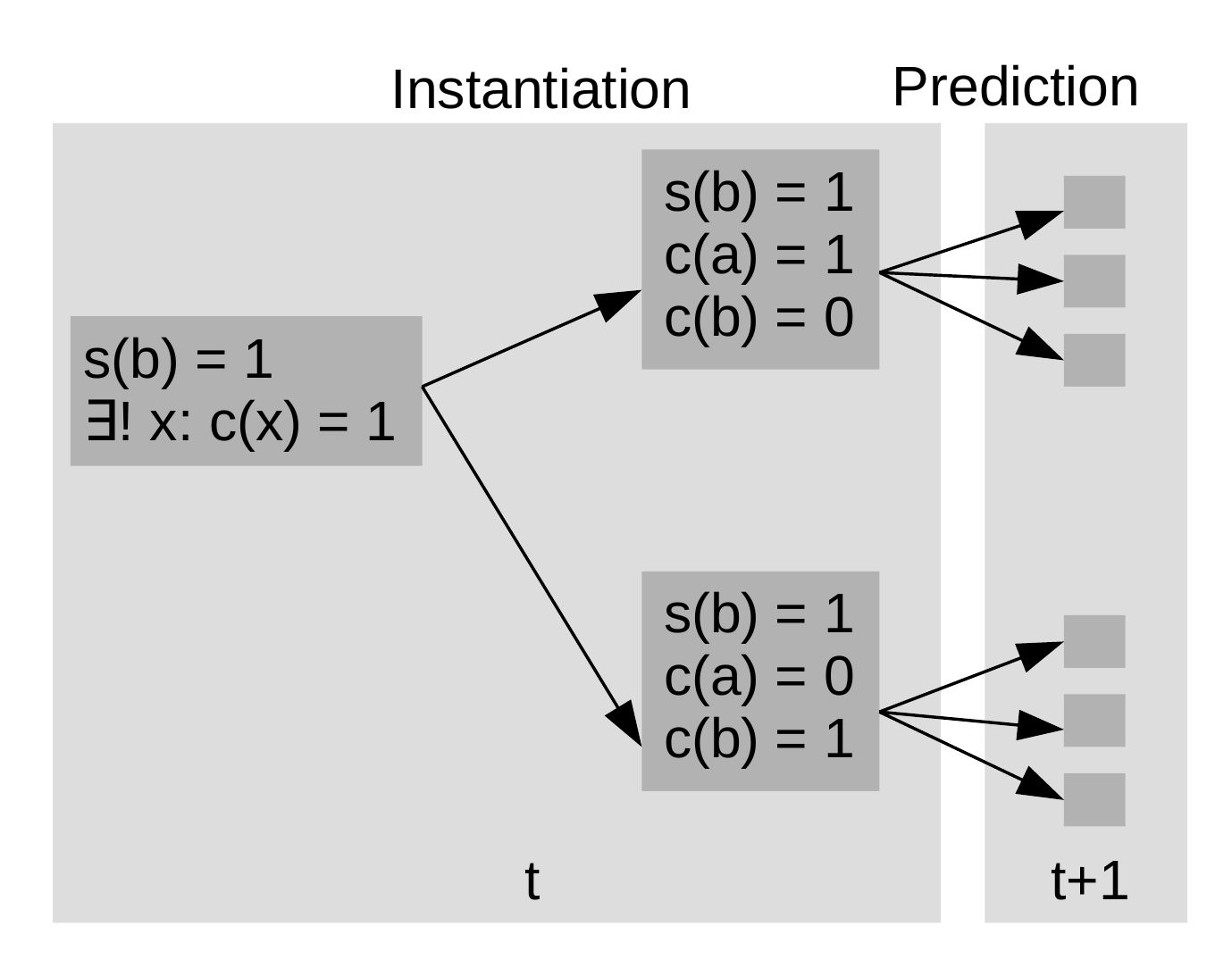}
\caption{Depiction of the logical particle filter for Example \ref{example:dynamicsmokers}. The instantiation step materializes all predicates necessary to calculate the transition model. Here, we assume that the values of $c(a)$ and $c(b)$ must be known to calculate the state transitions. Thus, all instantiations of $\exists ! x: c(x){=}1$ are materialized.}
\label{fig:lpf}
\end{figure}

The logical particle filter (LPF) \cite{zettlemoyer_logical_2008} is a Bayesian filtering algorithm where states are described by partially instantiated first-order logical formulae.
Each of those state descriptions actually represent a set of ground states (all instantiations of the formulae). 

\begin{example}
Consider the dynamic smokers scenario (Example \ref{example:dynamicsmokers}, Figure \ref{fig:smokers-dbn}).
Suppose we know that exactly one person has cancer, but we do not know which person. Furthermore, it is known that Bob smokes, and all other state variables are unknown. This situation can be represented by a single logical state in the LPF (representing the set of all 8 ground states that correspond to this situation):
\begin{equation*}
s(b) {=} 1, \exists ! x: c(x){=}1
\end{equation*}
Two examples for ground states that are represented by this logical state are:
\begin{equation*}
\begin{split}
s(a){=}1, s(b){=}0, c(a){=}1, c(b){=}0, d{=}0 \\
s(a){=}1,s(b){=}0, c(a){=}0, c(b){=}1, d{=}1
\end{split}
\end{equation*}
\end{example}
The transition model is described in terms of \emph{rules} that have preconditions and probabilistic effects.
 A state transition is performed as follows: First, a \textsf{split} operation is applied, which is necessary to determine which state transition rules are applicable in the current state. 

\begin{example}
\label{ex:smokers-lpf}
Suppose that the transition model requires that the \emph{specific} person having cancer is known (for example because the probability of Bob dying from cancer is higher than the probability of Alice dying from cancer).
The state
\begin{equation*}
s(b) {=} 1, \exists ! x: c(x){=}1
\end{equation*}
 is \textsf{split} into two states:
\begin{equation*}
\begin{split}
s(b) {=} 1, c(a) {=} 1, c(b) {=} 0\\
s(b) {=} 1, c(a) {=} 0, c(b) {=} 1
\end{split}
\end{equation*} 
  Note that these two states still represent multiple ground states each.
\end{example}
Afterwards, the transition model is applied to each state separately (in the same way as in a standard particle filter). 
The situation is depicted in Figure \ref{fig:lpf}. 

The LPF implicitly groups multiple RVs of a state: In the state $s(b) {=} 1, \exists ! x: c(x){=}1$, it is not specified \emph{which} specific person has cancer, only that the \emph{number} of people having cancer is one. In a way, this representation exploits the \emph{exchangeability} of the RVs $c(a)$ and $c(b)$ in the underlying distribution described by the state $\exists ! x: c(x){=}1$.
However, opposed to lifted inference algorithms, this capability to exploit exchangeability is limited: There is no formalism to specify that a \emph{specific number} of RVs have a certain value (like counting formulae in lifted inference), and no algorithmic solution to handle such cases has been proposed.

A problem not devised by the LPF is that predicates that are instantiated once stay instantiated for this particle, i.e.\ \textsf{merging} operations for LPFs have not yet been devised. This can lead to a complete propositionalization of the state space over time.
  \shortciteA{zettlemoyer_logical_2008} acknowledges that a \textsf{merging} operation would be necessary to apply LPF to realistic domains.

\subsection{Relational Particle Filter}
The relational particle filter (RPF) \shortcite{nitti_particle_2013,nitti_relational_2014,nitti_probabilistic_2016}  is a Bayesian filtering algorithm where states, as well as the transition and observation model, are described by \emph{distributional clauses}.
 
Distributional clauses are a  way to describe conditional probabilities, closely related to parfactors. They have the form $h \sim D \leftarrow B \sim= b$, which describes the probability $p(h|B{=}b)=D$. Each of $H$, $B$ and $D$ can have logical variables. For example, the clause
\begin{equation*}
\begin{split}
size(X) \sim beta(2, 3) \leftarrow material(X) \sim= metal
\end{split}
\end{equation*}
describes a conditional probability $p(size(X) \mymid material(X){=}metal)$ for each $X$. 
A \emph{dynamic} distributional clause (DDC) furthermore allows RVs to have time indices.  Thus, DDCs can be used to describe the conditional probabilities $p(x_{t}\mymid x_{t-1})$ and $p(y_t\mymid x_t)$ of Bayesian filtering models.
The algorithm performs particle filtering, using distributional clauses for the transition and observation model. Each particle is an assignment of values to the RVs, where some RVs may not have a specific value, but a distribution that is assigned to them. 

\begin{example}
Consider the dynamic smokers scenario (Example \ref{example:dynamicsmokers}, Figure \ref{fig:smokers-dbn}). 
The transition model is described in terms of a DDC. For example, the DDC
\begin{equation*}
\begin{split}
c(X)_t \sim bernoulli(0.5) \leftarrow s(X)_{t-1} \sim= 1 \\
c(X)_t \sim bernoulli(0.1) \leftarrow s(X)_{t-1} \sim= 0 
\end{split}
\end{equation*}
 describes that the probability of each person having cancer depends on the smoking state of this person at the previous time step. Other aspects of the transition and observation model are expressed in a similar fashion. 
As an example of a particle, suppose that one of the particles encodes the state where both persons do not smoke, but have cancer, and where the value of $d_t$ (whether at least one person died at time $t$) follows a Bernoulli distribution:
\begin{equation*}
s(a)_t{=}0, s(b)_t{=}0, c(a)_t{=}1, c(b)_t{=}1, d_t{\sim} bernoulli(0.1)
\end{equation*}
\end{example}

Thus, each particle actually describes a \emph{distribution} of ground states, similar to the Rao-Blackwellized particle filter (RBPF). For example, the state above describes a distribution of two ground states with $d=0$ and $d=1$. 
A transition might require to know the specific value of an RV. This is achieved by sampling from the corresponding distribution -- obtaining a new set of particles -- and applying the transition model to each particle separately. This procedure is an instance of \textsf{splitting}.
%
Similar to the LPF, the RPF can suffer from a complete grounding over time, as \textsf{merging} operations for the RPF have not yet been devised.

\subsection{Relational Kalman Filter}

\begin{figure}[t]
\centering
\includegraphics[scale=0.6]{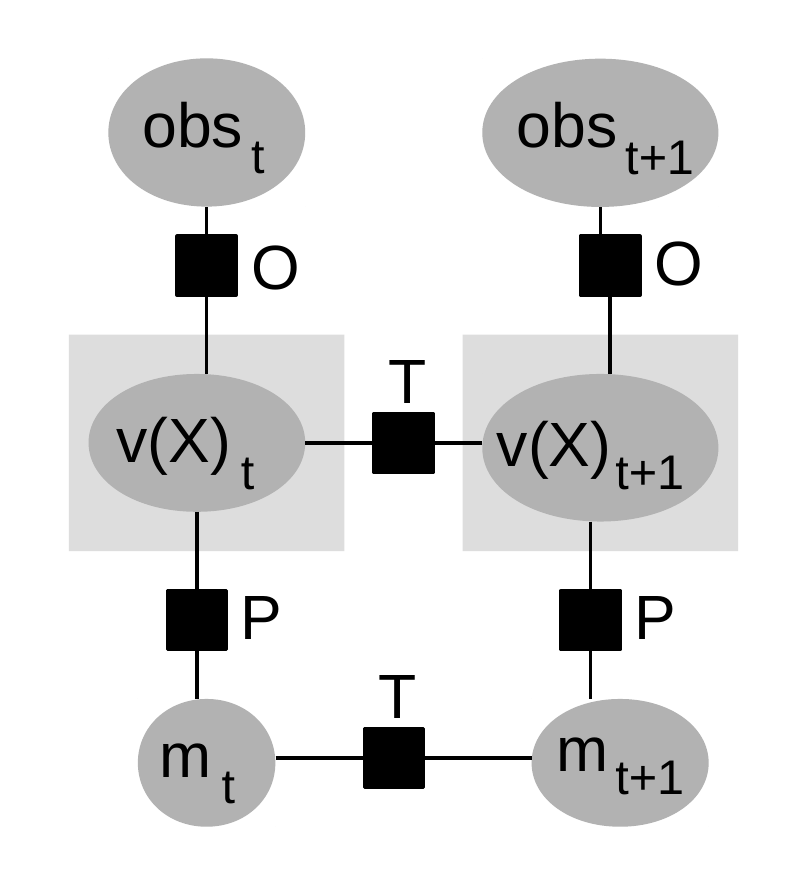}
\caption{Parfactor graph describing the relational Kalman filter for Example  \ref{example:rkf}. $P$ parfactors describe the state distribution, $T$ parfactors correspond to the transition model, $O$ parfactors correspond to the observation model.}
\label{fig:rkf}
\end{figure}

The relational Kalman filter \cite{choi_lifted_2011} is an algorithm for Bayesian filtering that is based on lifted inference, more specifically continuous FOVE \cite{choi_lifted_2010}.
The standard Kalman filter assumes a state that follows a multivariate normal distribution. 
 Opposed to that, the state of the system in the relational Kalman filter is modeled as a relational pairwise model (RPM) \shortcite{choi_lifted_2010}, an extension of parfactor graphs where the par-RVs are continuous and the parfactors are normal distributions of arity 2 (the latter is a technical condition, as the inference operations only work for these parfactors).
RPMs essentially represent a multivariate normal distribution with additional independence assumptions. 
The transition and observation model are also defined by RPMs. 
Based on this state representation, a Bayesian Filtering algorithm is defined, that is, \emph{predict} and \emph{update} steps are iteratively applied. Both steps are performed by employing continuous FOVE \shortcite{choi_lifted_2010}, i.e.\ by marginalizing out variables of the previous time step.

\begin{example}
\label{example:rkf}
The true value of a number of real estates is to be estimated over time, based on observations of sales prices and other factors, like the housing market index.
The value of real estate $i$ at time $t$ is modeled as a Gaussian RV $v_t(i)$, and the housing market index is modeled as a Gaussian RV $m_t$. At each step, several sales prices will be observed. If we initially assume each real estate to have an identical value, the estimated $v_t(i)$ will be the same for all unobserved $i$.
Thus, all of these values can be represented by a single, parametric RV $v_t(X)$.
The dependency between the state RVs at a single time step $t$ is represented by a parfactor (specifically, an RPM) $P\bigs{(}v_t(X),m_t\bigs{)}$ and the observation model is an RPM $O\bigs{(}v_t(X),obs_t\bigs{)}$. The transition model can (for example) be described by RPMs $T_v\bigs{(}v_t(X),v_{t+1}(X)\bigs{)}$ and $T_m\bigs{(}m_t,m_{t+1}\bigs{)}$. 
Figure \ref{fig:rkf} shows the parfactor graph describing the situation.
The predict and update steps thus have to be performed only once for each par-RV, instead of once for each RV.
For the predict step, the par-RVs $v(X)_t$ and $m_t$ are marginalized out of the joint distribution of par-RVs of time $t$ and $t+1$. For the update step, the distribution of par-RVs is updated, based on the new observation $obs_{t+1}$. 
\end{example}

The key challenge of the relational Kalman filter arises when individual observations about RVs corresponding to the same par-RV are made. In this case, in general, a \textsf{split} operation needs to be performed to handle each observed RV individually. Interestingly, splitting is not necessary when only the means of the ground RVs become distinct, but only when the variances of the RVs become distinct.
\citeA{choi_learning_2015} describe an algorithm to approximately \textsf{merge} variables that have become distinct due to observations.

This approach groups equivalent variables and reasons about them as a group (\textsf{group variables}), and also represents variables parametrically, as a Gaussian distribution (\textsf{pa\-ra\-me\-tri\-za\-tion}). Thus, it is the only approach we know of that exploits both types of lifting defined in this review.
However, the approach is limited in its applicability, because it only allows Gaussian RVs and a linear transition model.

\subsection{Data Association}
\label{subsec:data-ass}

\begin{figure}[t]
\centering
\includegraphics[scale=0.5]{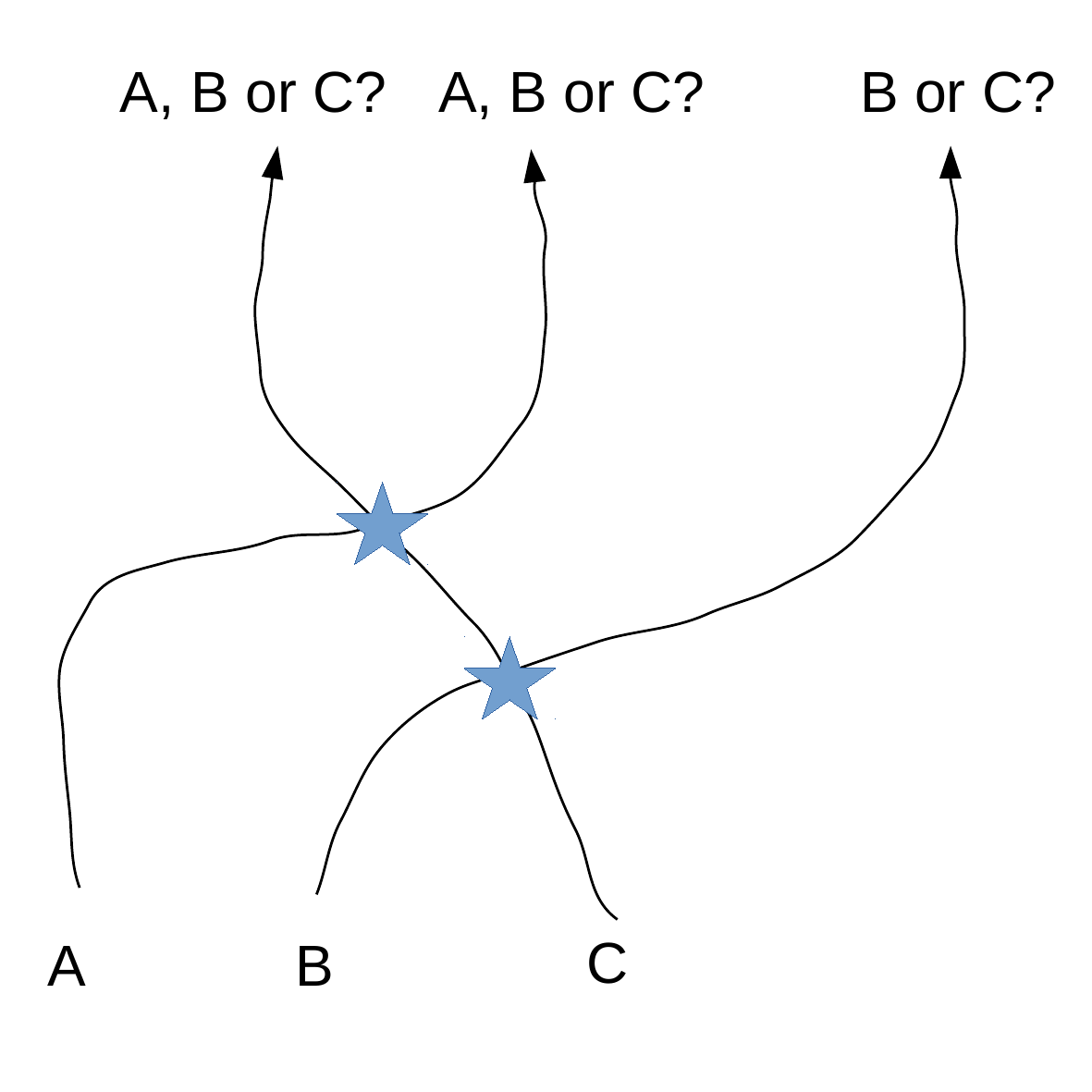}
\caption{Data association problem. Three objects A, B and C move in 2D space. The identities of the objects cannot be observed directly. When they come too close, we get confused about the correspondence of the objects and the tracks \shortcite<adapted from>{huang_fourier_2009}.}
\label{fig:data-association}
\end{figure}

Data Association algorithms are concerned with the following problem:
Given a number of \emph{tracks} $t_1,...,t_n$ (e.g. radar measurements, tracks of people in a video) that correspond to objects $o_1,...,o_n$, maintain the correct correspondence between tracks and objects (or, more general, a distribution of object-track associations). The problem is visualized in Figure \ref{fig:data-association}. This problem can be viewed as performing Bayesian filtering in a state space where each state is a permutation of objects. There are $n!$ many of these permutations, so the naive approach to maintain a distribution of those permutations explicitly suffers from the state space explosion problem. 
Thus, the central task of Data Association algorithms is to maintain an efficient representation of distributions of permutations, and mechanisms to perform the predict and update steps of Bayesian filtering directly on this representation.
Two conceptually different approaches for this goal have been devised.
The first one, known as the Fourier-theoretic approach \shortcite{huang_efficient_2009,huang_fourier_2009,huang_exploiting_2009,jagabathula_inferring_2011,kondor_multi-object_2007}, utilizes a Fourier transformation over the symmetric group $\mathbb{S}_n$ (the group that represents permutations of $n$ objects).
\new{
Instead of maintaining the complete distribution $p(\sigma)$, $\sigma \in \mathbb{S}_n$, the distribution is approximated by its first few Fourier matrices, just like a function $f(x)$, $x \in \mathcal{R}$ can be approximated by its first few Fourier coefficients. 
}

The second approach \cite{schumitsch_information-form_2005} maintains a compact representation of the distribution over permutations matrices by an \emph{information matrix} $\Omega$. The information matrix contains unnormalized marginal probabilities $\Omega_{ij}$ for each association of track $i$ with identity $j$.
The following example illustrates the approach. 

\begin{figure}
\begin{subfigure}{0.45\textwidth}
\begin{equation*}
\begin{pmatrix}
2 & 12 & 4 & 4 \\
1 & 2 & 11 & 0 \\
10 & 4 & 4 & 15 \\
5 & 2 & 1 & 2
\end{pmatrix}
\end{equation*}
\subcaption{Information Matrix.}
\label{fig:data-association-ex-1}
\end{subfigure}
\begin{subfigure}{0.45\textwidth}
\begin{equation*}
\hat{A} = \underset{A}{\text{argmax}}\ \text{tr}\, A^T \Omega = 
\begin{pmatrix}
0 & 1 & 0 & 0 \\
0 & 0 & 1 & 0 \\
0 & 0 & 0 & 1 \\
1 & 0 & 0 & 0
\end{pmatrix}
\end{equation*}
\subcaption{Most likely association.}
\label{fig:data-association-ex-2}
\end{subfigure}
\caption{Illustration of the information form approach for data association \shortcite<adapted from>{schumitsch_information-form_2005}.}
\label{fig:data-association-ex}
\end{figure}

\new{
\begin{example}
\label{ex:data-association-1}
Suppose we are tracking four objects. The distribution of object-track associations can be represented by the information matrix shown in Figure \ref{fig:data-association-ex-1}. The first column corresponds to track 1, and the values imply the association of this track with the four objects, suggesting that track 1 is most strongly associated with object 3 (since this is the largest value in the column).
However, the most likely permutation matrix, shown in Figure \ref{fig:data-association-ex-2}, shows that actually, track 1 is most likely associated with object 4 (i.e.\ it is not sufficient to consider the columns separately). 
\end{example}
}

\new{
Given the information matrix $\Omega$, we can calculate the probability of any permutation matrix $A$ as $p(A) = 1/Z\, \text{exp}\, \text{tr}\, A^T \Omega$. Calculating the partition function $Z$ is difficult, as it involves summing over all permutation matrices. 
However, the predict and update steps of the Bayesian filter can be performed directly on the information matrix: The observation of an association of a track $i$ with a specific object $j$ leads to an increase of the corresponding value $\Omega_{ij}$, and the mixing of tracks $i_1$ and $i_2$ leads to the same values in columns $i_1$ and $i_2$ in the information matrix.
}

Both approaches have been compared by \citeA{jiang_fourier-information_2011}.
They found that the Fourier-theoretic approach is better suited for scenarios with high uncertainty, while the information-theoretic approach is better suited for scenarios with low uncertainty about the data association. 
%

To sum up, both approaches represent a (high-dimensional) distribution compactly. They do this by transforming the distribution into a different space, where it is easy to find a compact, approximate representation. This transformation can be seen as a form of \textsf{parameterization}, as defined in Section \ref{sec:properties}: The Fourier coefficients are the parameters of a mixture model of complex exponential functions. 
Operations corresponding to \textsf{splitting} or \textsf{merging} are not necessary in this setting: The distribution is always represented in this transformed space, and a grounding (in this case, an transformation back into the original space) is never necessary.



%% file: figures/clustering-table.tex
\begin{tabular}{llllllrll}
  \toprule \rot{Online} & \rot{Identification} & \rot{Group Variables} & \rot{Parametrization} & \rot{Splitting} & \rot{Merging} & \rot{No. Papers} & \rot{Name} & \rot{Section} \\ 
  $\square$ & $\blacksquare$ & $\blacksquare$ & $\square$ & $\blacksquare$ & - & 50 & LI Top-down & 6.1.1 \\ 
  $\square$ & $\blacksquare$ & $\blacksquare$ & $\square$ & - & $\blacksquare$ & 31 & LI Bottom-up & 6.1.2 \\ 
   $\square$ & $\blacksquare$ & $\square$ & $\blacksquare$ & $\blacksquare$ & $\blacksquare$ & 5 & Continuous Inference & 6.2 \\ 
  $\blacksquare$ & $\square$ & $\blacksquare$ & $\square$ & - & - & 7 & Multiset Rewriting & 6.3 \\ 
  $\blacksquare$ & $\blacksquare$ & $\blacksquare$ & $\square$ & $\blacksquare$ & $\square$ & 1 & Logical Particle Filter & 6.4\\ 
  $\blacksquare$ & $\blacksquare$ & $\square$ & $\blacksquare$ & $\blacksquare$ & $\square$ & 3 & Relational Particle Filter & 6.5 \\ 
  $\blacksquare$ & $\blacksquare$ & $\blacksquare$ & $\blacksquare$ & $\blacksquare$ & $\blacksquare$ & 3 & Relational Kalman Filter & 6.6 \\ 
  $\blacksquare$ & $\blacksquare$ & $\square$ & $\blacksquare$ & - & - & 16 & Data Association & 6.7 \\ 
   \bottomrule \end{tabular}

%% file: Input/06_Future_Research.tex

\section{A Guide to Identify Suitable State-Space Abstraction Approaches}
\label{sec:application-view}
The goal of this section is to provide a useful guideline for practitioners to identify appropriate algorithms (or algorithmic ideas) for a given problem. It also summarizes our findings regarding research question \textbf{Q2}: \emph{How can we characterize the problem classes that each of the 8 groups of approaches can solve?}  We do so by rephrasing the properties of the algorithms as properties of the problem domains. 
At the same time, this perspective shows interesting problem domains that are not addressed by current approaches, and therefore identifies interesting directions for future research. 
As stated previously, the properties of the algorithms (see Table \ref{table:clustering}) directly provide a characterization of the application domain:
\begin{itemize}
\item \textsf{Online} algorithms are applicable to inference problems for sequential processes (e.g. the dynamic smokers domain of Example \ref{example:dynamicsmokers}).
\item \textsf{Identification} is necessary in two cases: Either observations about individuals are made (e.g. we observe that an individual person smokes), or the individuals need to be distinguished for some other reason (for example, because the transition model in a dynamic model requires to know the value of an individual RV, as in the variant of the dynamic smokers domain in Example \ref{ex:smokers-lpf}).
\item \textsf{Grouping of Variables} means that the algorithm can exploit exchangeability in the state space, i.e.\ a regular structure between multiple variables. Algorithm that have this capability can potentially solve some problems (that exhibit exchangeability) more efficiently (typically, for problems that do do not exhibit exchangeability, the algorithms simply resort to propositional inference). For example, algorithms that can group variables can potentially solve the smokers domain (Example \ref{ex:smokersbn}) more efficiently that propositional inference algorithms.
\item \textsf{Parametrization} allows the inference algorithm to exploit a regular structure in the distribution of a single variable. This is necessary for domains with continuous variables (see Example \ref{ex:continuous-inference}), but discrete domains (like Data Association tasks) can also benefit from parametrization.
\item \textsf{Splitting} operations are necessary for algorithms that start with a lifted representation (for example, a lifted graphical model or a first-order logic state representation, as in the Logical Particle Filter) and then need to identify individuals (as outlined for the \textsf{identification} property).
\item \textsf{Merging} operations make the algorithm applicable to problems that are given in propositional form, but still contain symmetric properties (e.g. the ground factor graph for the smokers domain, Example \ref{ex:smokersbn}). Merging is also useful in cases where the representation propositionalizes over time due to repeated splitting: In these cases, merging operations can re-introduce a compact representation (as for example done in the relational Kalman filter,  \shortciteR<see>{choi_learning_2015}). 
\end{itemize}
\begin{figure}[t]
\centering
\includegraphics[scale=0.35]{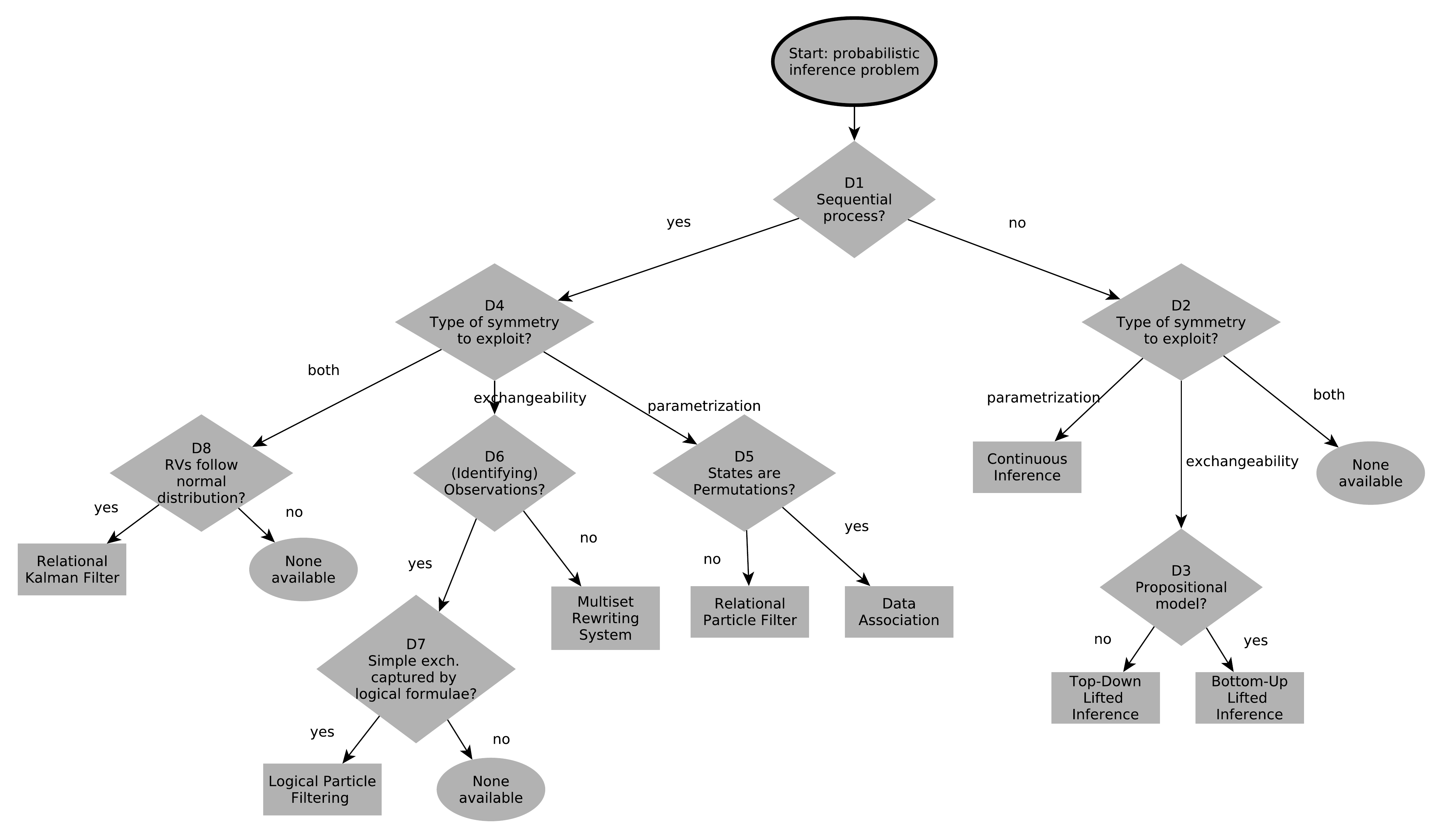}
\caption{Decision diagram to decide on appropriate method to solve a given problem instance. Diamonds denote decisions, rectangles denote categories of inference algorithms, and paths to ellipses describe problems for which no inference algorithm (that performs state-space abstractions) exists yet.}
\label{fig:flowchart-problems}
\end{figure}
Based on these considerations, we can identify approaches or algorithmic ideas that are suitable to solve a given problem in a lifted way (see Figure \ref{fig:flowchart-problems}).
The first decision (\textbf{D1}) is concerned with whether the system develops sequentially over time, constantly receiving new observations (requiring an approach capable of \textsf{online} inference), or not.
Next, one needs to decide on the type of state space abstraction that the inference problem is susceptible of (\textbf{D2} and \textbf{D4}): Either the distribution exhibits exchangeability, or some (marginal) distribution can be represented in a parametric form, or both (if both is not the case, then the inference problem does not allow any state space abstraction of the types investigated in this review). 
For non-sequential problems (\textbf{D2}), continuous inference algorithms can be used when the probabilistic model contains continuous variables whose distributions can be modeled (or approximated) by piecewise polynomial functions.
When some of the RVs in the model are exchangeable, lifted inference algorithms can exploit this fact for efficient inference. 
Depending on the input format of the model (\textbf{D3}), bottom-up or top-down lifted inference algorithms can be applied. For lifted inference to be polynomial in the domain size, certain conditions have to apply, as discussed in Appendix \ref{sec:li-complexity-classes}. However, even when these conditions do not apply, lifted inference can be more efficient than propositional inference.
A combination of continuous and lifted inference algorithms, that can exploit both exchangeability and parametric distributions, has not been devised yet.

Sequential processes, on the other hand, require online algorithms to cope with indefinite observation sequences and unlimited numbers of RVs. 
Here again, one has to decide on the applicable type of state space abstraction (\textbf{D4}). When some of the marginal distributions can be represented in parametric form, the relational particle filter (or Rao-Blackwellized filtering in general) can be used. 
Some specific form of parametrization -- data association methods -- can be used when the state space consists of permutations (\textbf{D5}). 

There are two categories of online algorithms that exploit exchangeability: When no observations are made (\textbf{D6}), Multiset rewriting systems can be used. Otherwise, the logical particle filter may be applicable, which has, however, very limited capabilities to make use of exchangeability (\textbf{D7}). 
The relational Kalman filter exploits exchangeability, and also makes use of parametric distributions -- as long as all distributions are Gaussian (\textbf{D8}).

Finally, coming back to the three examples from the introduction, Example \ref{example:smokers} (smokers) can be solved with a lifted inference algorithm, Example \ref{example:tracking} (office) with a data association approach, and Example \ref{example:chemistry} (biochemical reaction) with a multiset rewriting system.

\section{Conclusion and Future Work}
\label{sec:futurework}
In this section, we discuss the open problems identified in Section \ref{sec:application-view} in more detail. We propose ideas on how the methods identified in this review could be combined or extended appropriately, to devise an algorithm that can solve this problem class.

\subsection{Future Work}
\label{subsec:futurework}
As becomes obvious from the flow chart in Figure \ref{fig:flowchart-problems}, there is no algorithm that can exploit exchangeability (as lifted inference does) \emph{and at the same time} handle (continuous) distributions parametrically (as continuous inference algorithms do). One can easily imagine a scenario where this would be beneficial: Consider an object localization task similar to Example \ref{ex:continuous-inference}, but multiple objects are observed.
A combination of both state space abstraction types is, for example, conceivable for variable elimination, for which both lifted inference \cite{taghipour_generalized_2014} and parametric inference \cite{sanner_symbolic_2012-1} approaches exist, or weighted model counting, for which also both lifted inference \shortcite{van_den_broeck_lifted_2011} and parametric inference \shortcite{belle_probabilistic_2015} exists.

For sequential inference tasks, there is no algorithm that can exploit exchangeability to the same extend as lifted inference does for non-sequential inference. This would, however, be relevant for sequential inference problems exhibiting exchangeability, like the dynamic smokers domain presented in Example \ref{example:dynamicsmokers}. The relational Kalman filter requires all factors to be Gaussian, and logical particle filtering cannot handle statements about counts, like ``exactly 3 RVs of this group of RVs are true, and it does not matter which''. Multiset rewriting systems (MRSs) can efficiently handle systems with exchangeable RVs -- however, current MRSs do not provide a mechanism to incorporate evidence about specific individuals (say, we know that Bob is having cancer at $t=10$).  
In the following, we sketch two ideas that seem promising for developing a general lifted Bayesian filtering algorithm.

One idea is to base such an algorithm on an MRS, i.e.\ a Bayesian filtering algorithm with a multiset-based state description and a transition model defined in terms of rewriting rules.
Such a system directly allows to group equivalent aspects of the state by the multiset state representation. The crucial aspect for such a system is the way the state space abstractions are represented, i.e.\ how similar entities can be grouped, despite the fact that they may not be completely the same (e.g. because we have distinct observations about them).

The other idea is to base the algorithm on lifted inference approaches, and examine how they can be used to implement the predict and update step of Bayesian filtering. A first step in this direction are filtering algorithms for dynamic MLNs \cite{geier_approximate_2011,papai_slice_2012}, that require using a probabilistic inference algorithm at each time step. However, the effects of using a \emph{lifted} inference algorithm each time has not been evaluated yet, and it is unclear how to maintain a lifted state representation over time.

Furthermore, exploiting both exchangeability and parametric distributions is also relevant for sequential inference algorithms. As an example, consider the localization task from Example \ref{ex:continuous-inference}, but for multiple agents instead of a single agent. Here, some RVs (locations, sensor measurements) are continuous, and RVs of different agents can be exchangeable (e.g. we might know that two agents have the same location distribution, one agent has a different location distribution, but we do not know which agent is associated with which distribution).

In general, one of the most challenging aspects of inference algorithms that exploit symmetries is the question how to prevent the state representation from degenerating (become increasingly grounded), as individual observations that break the symmetries in the state space are received. 
\shortciteA{kersting_lifted_2012} notes:
``Even if there are symmetries within a probabilistic model, they easily
break when it comes to inference since variables become correlated
by virtue of depending asymmetrically on evidence'' (p.~37).
This is specifically problematic for \emph{exact} algorithms, that need to consider even slight symmetry breaks. 
For non-sequential models, approaches that can handle this problem by finding \emph{approximate} symmetries in the model have been proposed \shortcite{singla_approximate_2014,venugopal_evidence-based_2014,van_den_broeck_lifted_2015} -- which can gain even more efficiently by also using approximate inference algorithms (like belief propagation or sampling).

The problem is even more prevalent in sequential lifted inference algorithms: Even when evidence for a single time step leads only to a slight degeneration of the symmetries, over time, the complete model will become ground.
On the other hand, the prediction step in Bayesian filtering might lead to an \emph{increase} in symmetry: The intuition here is that the prediction in general increases the uncertainty of the state estimate, potentially (partially) revoking the effect of symmetry-breaking evidence. 
There is only very few research on this aspect for sequential models. 
For the relational Kalman filter, an approach has been devised that approximately regroups state variables after symmetry-breaking evidence has been observed \shortcite{choi_learning_2015}, relying on the fact that the difference in these variables can be bound under certain conditions. 
In general, the idea of occasionally performing operations that re-introduce (approximate) symmetries seems to be a promising idea for more general lifted Bayesian filtering algorithms.

\subsection{Conclusion}
\label{subsec:conclusion}
Probabilistic inference is the task to derive the probability of certain random variables, given the values of other variables and a model for the relationship between the variables.
In many cases, symmetries and redundancies are implicitly present in the model, which cannot be exploited by conventional inference algorithms.
In the last 15 years, inference methods have been devised that can exploit the symmetric structure to speed up inference and thus make it feasible for much larger models.

In this article, we presented the results of a systematic review concerned with these methods. 
We identified eight classes of such inference algorithms, which have been grouped based on their common properties, and thus the common problems they can be applied to.
For the first time, this systematic review presented a unified view of these methods, that  have been devised by different research communities. 
Specifically, we emphasized inference algorithms for sequential processes (Bayesian filtering), a relevant application domain that has been neglected by lifted inference algorithms, and is not discussed in previous reviews of the same topic \shortcite{kersting_lifted_2012,kimmig_lifted_2015}. 
We found that no Bayesian filtering algorithm has been devised yet that can exploit symmetries to the same extend as lifted inference algorithms do for non-sequential inference.
Developing such an algorithm might be approached by employing ideas from lifted inference or multiset rewriting systems.

One of the main problems underlying all approaches is \emph{symmetry-breaking evidence} that makes it difficult to maintain a lifted representation. This problem is very prevalent in real-world scenarios, like sensor data processing, and solutions for non-sequential inference algorithms based on finding approximate symmetries have been proposed.
Investigating how to cope with this problem in the context of Bayesian filtering is an interesting future research topic.

%% file: Input/07_AppendixDomainLifted.tex

\section{Lifted Inference Complexity Classes}
\label{sec:li-complexity-classes}
\new{
Recently, attempts have been made to structure the problem classes for lifted inference algorithms, based on whether they can be solved efficiently.
In general, there is no guarantee that lifted inference is tractable (i.e.\ has a polynomial runtime),  \shortciteA{jaeger_liftability_2012} even showed the existence of intractable inference problems. 
}

\new{
However, there are problem classes for which tractability guarantees can be given. 
To analyze them, it is useful to define inference problems in terms of \emph{weighted model counting} on a first-order knowledge base (see Section \ref{subsubsec:top-down-li}). 
%
Using this representation, different problem classes can be defined regarding the specific fragment of first-order logic used ($FO$: function-free first-order logic and $RFOL$: $FO$ without constant symbols), allowed quantifiers, and the maximum number of logical variables per formula.
}

\new{
The central notion is that of \emph{domain-lifted} algorithms. An algorithm is domain-lifted for a problem class, iff for all instances of this problem class, inference is polynomial in the domain size of the logical variables. 
Table \ref{table:liftability} shows domain-liftability results for different algorithms and problem classes.
Note that this table shows only results regarding domain-lifting. Results regarding other definitions of lifting (e.g. approximate liftability) are discussed by \shortciteA{jaeger_liftability_2012}.
}

\new{
It turns out that inference on knowledge bases with at most two logical variables per formula ($FO^2$) is domain-liftable, i.e.\ all instances of this class can be solved in polynomial time with respect to the domain size (as well as a generalization, $S^2FO^2$). Furthermore, at least two inference algorithms are known that can actually perform inference for this problem class in polynomial time: WFOMC, as proposed by \shortciteA{van_den_broeck_completeness_2011} and \shortciteA{van_den_broeck_skolemization_2014}, and the FOVE variant of \shortciteA{taghipour_completeness_2013-1}.
On the other hand, it was shown that for general $FO$, Lifted Inference is not polynomial in the domain size.
}

Another class of inference problems that is known to be domain-lifted is \emph{recursive unary} ($RU$), which basically describes the problems that can be solved in polynomial time by lifted recursive conditioning (Section \ref{subsubsec:top-down-li}): A theory is in RU when exhaustively applying the rules of lifted recursive conditioning leads to a theory which contains a predicate that has only a single logical variable. Then, the algorithm can branch on this atom, generating branches for all \emph{numbers} of corresponding RVs being true. 
It has been shown that $RU$ subsumes $FO^2$ \shortcite{kazemi_new_2016}.

\new{
An example of a problem where no domain-lifted marginal inference algorithm is known is the \emph{transitive} formula $friends(X,Y) \land friends(Y,Z) \Rightarrow friends(X,Z)$ (although MAP inference can be performed efficiently for this formula, \citeR<see>{mittal_new_2014}).
}

\begin{table}[tb]
\begin{center}
\leavevmode
\begin{tabular}{p{5cm}llp{5cm}}
\toprule
Algorithm & KB & DL  & Reference \\
\midrule
All   & $RFOL$($\forall \exists =$) & $\square$ &  \shortcite{jaeger_complexity_2000-1} \\
WFOMC \shortcite{van_den_broeck_lifted_2011,van_den_broeck_skolemization_2014} & $FO^2$($\forall \exists =$)  & $\blacksquare$ &  \shortcite{van_den_broeck_completeness_2011,van_den_broeck_skolemization_2014} \\
WFOMC  \shortcite{beame_symmetric_2015} &$\gamma$-acyclic query\footnote{\shortcite{fagin_degrees_1983}} & $\blacksquare$ &  \shortcite{beame_symmetric_2015} \\
\midrule
LRC  \shortcite{poole_towards_2011} & $RU$ & $\blacksquare$&  \shortcite{poole_towards_2011} \\
LRC  \shortcite{kazemi_new_2016} & $S^2FO^2$(=)\footnote{This class is similar to $FO^2$(=), except that it may contain additional clauses that use a single binary predicate $S$ such that each clause has exactly two different literals of $S$. }  & $\blacksquare$&  \shortcite{kazemi_new_2016} \\
LRC  \shortcite{kazemi_new_2016} & $S^2RU$ & $\blacksquare$&  \shortcite{kazemi_new_2016} \\

\midrule
FOVE  \shortcite{de_salvo_braz_lifted_2005} & $FO^1$(=)  & $\square$ &  \shortcite{taghipour_completeness_2013-1} \\
C-FOVE  \shortcite{milch_lifted_2008} & $FO^1$(=)  & $\square$ &  \shortcite{taghipour_completeness_2013-1} \\
C-FOVE$^\#$  \shortcite{taghipour_generalized_2014,apsel_extended_2011} & $FO^1$(=)   & $\blacksquare$ &  \shortcite{taghipour_completeness_2013-1} \\
\midrule
C-FOVE$^\#$  \shortcite{taghipour_generalized_2014,apsel_extended_2011} & $FO^2$(=)   & $\square$  & \shortcite{taghipour_completeness_2013-1} \\
C-FOVE$^+$  \shortcite{taghipour_completeness_2013-1} & $FO^2$(=)  & $\blacksquare$ &  \shortcite{taghipour_completeness_2013-1} \\
\bottomrule
\end{tabular}
\caption{Liftability results for different algorithms and problem classes. ``All'' algorithms mean that the result applies to all lifted inference in general. KB: Knowledge Base, DL: Domain-lifted, $\square$: Not domain-lifted, $\blacksquare$: Domain-lifted \shortcite<adapted from>{taghipour_completeness_2013,jaeger_liftability_2012}.}
\label{table:liftability}
\end{center}
\end{table}

%% file: Input/07_AppendixB.tex

\section{Related Approaches}
\label{subsec:related}
There are multiple approaches that touch upon related topics as the ones explored in this review, but have not been included.
In the following, we will discuss their connection to the approaches examined by this review, and  argue why each of them did not match our inclusion criteria. 

\subsection{Knowledge-Based Model Construction}
Knowledge-based model construction (KBMC) is a type of inference algorithm for lifted graphical models. They work by completely grounding the model and performing standard probabilistic inference in the propositional model. 

There are numerous extensions and improvements that have been proposed for these algorithms.
For example, \shortciteA{richardson_markov_2006} ground only those formulae necessary to answer the query. \citeA{singla_memory-efficient_2006-2} propose a \emph{lazy} KBMC algorithm that performs grounding on the fly.
\citeA{glass_focused_2012} propose an approximate algorithm that only produces the most relevant ground formulae, and ignores the rest.
Using these methods, KBMC approaches can be more efficient than standard, propositional inference. However, at their core, they perform propositional inference.
Thus, they do not match inclusion criterion 8.

\subsection{Knowledge Compilation and Arithmetic Circuits}
\label{subsec:knowledge-comp}

Arithmetic circuits are data structures that compactly represent functions. Early approaches like Binary Decision Diagrams (BDDs) represent boolean functions, but extensions have been devised for representing functions of the type $\mathbb{B}^n \rightarrow \mathbb{R}$.
\citeA{darwiche_knowledge_2002} provide a detailed comparison of such approaches. 
The appeal of these methods is that they allow to efficiently answer specific classes of queries. 
\emph{Knowledge compilation} exploits this fact, by tranforming logical formulae into such a formalism, which then allows efficient inference. The motivation is to perform this (potentially costly) transformation up-front, and then be able to answer a large number of queries on the compiled representation very fast.

This idea can be also be used for \emph{probabilistic} inference: As discussed in Section \ref{subsubsec:top-down-li}, probabilistic inference can be transformed into a weighted model counting (WMC) problem. Knowledge compilation can then be used to solve the WMC problem efficiently. 
On the other hand, arithmetic circuits can also be used directly to represent distributions (instead of using a conventional graphical model). Examples include variants of Ordered Binary Decision Diagrams \cite{jaeger_probabilistic_2004,dal2017weighted}, Algebraic Decision Diagrams \cite{sanner_affine_2005}, and Sum-Product Networks \cite{gens_learning_2013}.
In these formalisms, some probabilistic inference operations can be performed in polynomial time (in the size of the circuit).  


These approaches have not been included in this review because they do not perform any of the two investigated state space abstraction methods: They do not group similar RVs, and they do not operate on parameters of the distributions (i.e.\ they do not match the definition of state space abstraction that we use). 
Still, they are related to the methods presented here insofar as they pursue the same overall goal: Compact representations of distributions, and efficient operations on this representations that lead to efficient inference. 

Knowledge compilation methods and arithmetic circuits have successfully been combined with other state space abstraction methods: Combining knowledge compilation with lifted inference ideas yields \emph{first-order} knowledge compilation \shortcite{van_den_broeck_lifted_2011}, discussed in Section \ref{subsubsec:top-down-li}. 
Arithmetic circuits (specifically, ADDs) are used for symbolic variable elimination (outlined in Section \ref{subsec:cont-inference}), to keep the representation of the case statements compact.

\subsection{Markov Decision Processes}
A Markov decision process (MDP) is a model for sequential decision making where an agent has to select actions based on the current environment state. Each action is associated with a reward.
Given an MDP, the task is to compute an optimal \emph{policy}, i.e.\ a function that assigns each state a corresponding action such that the long-term reward is maximized.
The optimal policy can be obtained by computing the \emph{value function} (that assigns a value to each state) using dynamic programming. \citeA{puterman_markov_2014} provides a more thorough introduction into algorithms for solving MDPs.

MDPs also suffer from the state space explosion problem, and solutions similar to some of the algorithms discussed in Section \ref{sec:analysis} have been developed. These methods follow two basic ideas.
The first approach is to find symmetries in the state space of an MDP and group symmetric state, thus obtaining a smaller state space \cite{dean_model_1997,givan_equivalence_2003,kang_exploiting_2012}.
The second approach is to perform all operations within a more compact first-order representation \cite{boutilier_symbolic_2001-1,kersting_bellman_2004-1,holldobler_logic-based_2004,sanner_practical_2009-1,wang_first_2008-1}.
In these approaches, states, actions, reward functions, value functions and policies are all based on   first-order logic. This way, the resulting policy can be independent of the actual domain objects, and the computations to obtain this policy can be independent of the domain size. 
A problem is that the (logical) representation of the value function can easily become very complex and redundant, requiring expensive first-order simplification. 
A first-order extension of ADDs (first-order ADDs, FOADDs) has been devised, which can be used to compactly represent the value function. 
Furthermore, approximate methods (like approximate first-order linear programming), that avoid this problem, can be used. 
Conceptually, first-order MDPs (and their solution techniques) bear strong relationships to lifted probabilistic inference: Both are concerned with first-order models, where parts of the model are redundant or identical. They both exploit these symmetries to achieve more efficient algorithms, by performing operations ``in bulk'' for entire sets of redundant components. 

However, there is also a more technical, intimate relationship between MDPs and probabilistic inference: It has been shown that decision problems (in terms of an MDP) can be cast into a probabilistic inference problem \cite{toussaint_probabilistic_2006}. Thus, any probabilistic inference algorithm can be used to solve MDPs. 
From this point of view, first-order MDPs are an \emph{application domain} of probabilistic inference (although research on both topics has mostly been distinct).
This relationship also holds for first-order MDPs: 
Recently, \citeA{khardon_stochastic_2017} showed that the probabilistic inference problem that can be derived from a first-order MDP inherits its symmetric structure.
This structure can be exploited by lifted inference, avoiding redundant computations.
Due to the complex structure of the query, it is however not possible to use standard lifted inference algorithms here. Instead, the first-order dynamic programming approaches can be seen as performing some specialized lifted inference algorithm (that is completely independent of the domain size).  
An interesting perspective for future research is to combine the distinct innovations from both domains, and close the gap between the respective lines of research -- a paradigm termed \emph{generalized lifted inference} by \shortciteA{khardon_stochastic_2017}.

Nevertheless, we chose to not consider first-order MDPs in Section \ref{sec:analysis} of this review. The reason is that the dynamic programming-based algorithms that are at the heart of solving MDPs do not \emph{directly} involve probabilistic inference, and thus some of the properties derived in Section \ref{sec:properties} are not meaningful for these algorithms. In other words, dynamic programming does not match inclusion criterion 6. 


\subsection{Statistical Relational Learning}

A relevant question not discussed so far in this review is that of \emph{learning} first-order probabilistic models. So far, we assumed that the models are given, and the only task is perform inference in these models. For many application domains, learning the model is one of the most relevant (and most challenging) aspects. 
For example, in tasks like link prediction (decide whether a specific relation exists between two objects) or entity resolution (decide which records in a database refer to the same real-world entity), we are given a rich, relational structure, and want to estimate a first-order probabilistic model describing this structure (to then perform inference in this model). 
The research field investigating this task is known as \emph{statistical relational learning}. For an overview of the methods used in this field, we refer to the book by \shortciteA{getoor2007introduction}. 

One can distinguish \emph{parameter learning}, where the structure of the probabilistic model is given, and \emph{structure learning}, where even the structure needs to be learned. 
In parameter learning, the goal is to optimize the likelihood of the model, given the data. This is, as in the propositional setting, typically done by \emph{Expectation Maximization}: The parameters are computed in an iterative process, consisting of computing the expected likelihood of the model, given the current parameters, and maximizing this expectation function. In contrast to propositional models, however, multiple parameters may be tied in relational models (thus effectively reducing the total number of parameters). Parameter learning is a difficult task, as it requires to perform probabilistic inference (which is itself a hard problem) each time the expectation is computed. Thus, approximate methods are typically used, that optimize easier to compute measures than the likelihood. 
Recently, exact \cite{van_haaren_lifted_2016} and approximate \cite{ahmadi_exploiting_2013} lifted inference has been used for parameter learning. 
Structure learning is even more challenging: The structure is also learned in an iterative process, requiring parameter learning at each step. 
Learning methods have been devised for a large number of probabilistic relational formalisms, including MLNs \shortcite{richardson_markov_2006,khot_learning_2011}, Problog \cite{gutmann_learning_2011}, CP-logic \cite{thon_stochastic_2011}, PRISM \cite{sato_parameter_2001}, probabilistic relational models \cite{getoor_learning_2002} and  Bayesian logic  programs \cite{kersting_adaptive_2001-1}. 

As this paper focuses on \emph{inference} rather than learning, these methods are not discussed in the main part of this review (i.e.\ they do not match inclusion criterion 6). 

\subsection{Logical Hidden Markov Models}
Logical Hidden Markov Models (LHMMs) \cite{kersting_logical_2006,natarajan_logical_2008,yue_filtering_2015,yue_logical_2015} are similar to Hidden Markov Models (HMMs), except that each state consists of a logical atom.
A LHMM transition consists of two steps. First, a ground atom is sampled based on the current state, i.e. the current logical atom. Then, an abstract transition is selected whose precondition matches the ground atom. This transition leads to a new abstract state. 
The filtering algorithm that has been presented for this representation requires considering all ground atoms. Thus, this approach does not match inclusion criterion 8.

\subsection{Probabilistic Model Checking}
Model Checking is concerned with the following problem: Given an abstract system specification, test if certain properties (defined in a temporal logic like LTL or CTL logic) are satisfied by the system. 
These specifications define a state space that is exhaustively searched to verify the property. 
A common technique is to not represent the state space explicitly, but \emph{symbolically} as a propositional formula, that in turn is represented as a binary decision diagram (BDD).
\emph{Probabilistic} model checking furthermore models state transition probabilities.

In Model Checking, the state space explosion problem is very common. For example, when the system consists of multiple concurrent processes, each execution ordering needs to be considered, which leads to a combinatorial explosion in the state space \cite{clarke_progress_2001}. 
Symbolic state space representation is one way to handle this problem. When the state space has a certain regular structure, the BDD representation can be much smaller than representing the state space explicitly.
Other methods directly reduce the number of states, the most prominent ones being 
 partial order reduction (POR) \cite{valmari_stubborn_1989,peled_all_1993,godefroid_partial-order_1996} and symmetry reduction \cite{clarke_symmetry_1998}.
These reduction methods follow similar ideas than bottom-up lifted inference algorithms: Starting with a propositional model, and finding symmetries in this model. 
Then, the model can be represented by a single representative of each set of symmetric state.

The reasons for excluding these approaches are similar to the reasons for excluding MDP-based approaches: 
Although they contain interesting ideas for state space reduction, the task and the used algorithms are completely different. 
This also means that the type of symmetry considered is quite different: In lifted inference, the symmetries must preserve the (conditional) probabilities of the RVs. In model checking, the symmetries must preserve the property we want to check.

\subsection{Multiple Hypotheses Tracking}
There is a large number of papers from the \emph{data association} community that have not been included in this review.
 A prominent example for this class of algorithms is the multiple hypotheses tracker \cite{reid_algorithm_1979}. It maintains all possible associations of measurements to objects explicitly.
 Therefore, it suffers from the state space explosion problem. 
 Several approximation methods, like pruning (keeping only the most likely hypotheses) \cite{cox_efficient_1996} have been developed.
 Other data association approaches have been proposed by \citeA{fortmann_sonar_1983,han_algorithm_2004}, and \citeA{oh_markov_2004}.
None of these approaches employ state space abstractions, which is the reason why we did not consider them for this review.

\subsection{Probabilistic Situation Calculus}
The situation calculus \cite{reiter_frame_1991} is a first-order logic formalism to reason about dynamic domains that are changed by actions. 
Several approaches combine the situation calculus with some form of probabilistic model.
In the works of \citeA{mateus_probabilistic_2001}, and \citeA{hajishirzi_sampling_2008}, actions have probabilistic effects. \citeA{bacchus_reasoning_1995,bacchus_reasoning_1999}, and \shortciteA{mateus_observations_2002} introduce uncertain observations (uncertainty about the current state). 
 The problem that is solved by these approaches is: \emph{Given a sequence of actions and an initial state, what is the probability that a first-order formula is true in the final state, after executing these actions?}
This is done by providing an explicit distribution over all possible states \shortcite{bacchus_reasoning_1995,bacchus_reasoning_1999}, or by sampling-based approaches \shortcite{mateus_probabilistic_2001,mateus_observations_2002,hajishirzi_sampling_2008}.

This formalism provides a compact state representation, by representing states using first-order logic. 
However, no algorithm that can reason efficiently in this representation has been devised. 
In fact, the state representation can become arbitrarily complex, as noted by \citeA{boutilier_symbolic_2001-1}.

%% file: Input/07_AppendixC.tex

\section{Assignment of Papers to Groups}
\label{sec:paperstogroups}
The following table shows the specific papers associated with each of the groups defined in Section \ref{sec:results}.

\input{figures/clustering-table-allcites.tex}

%% file: figures/clustering-table-allcites.tex
\begin{longtable}{lp{10cm}}
  \toprule Name & References \\ 
  \midrule  
 Top-down LI  & \shortcite{poole_first-order_2003} \shortcite{kisynski_constraint_2009} \shortcite{de_salvo_braz_lifted_2005} \shortcite{de_salvo_braz_mpe_2006} \shortcite{milch_lifted_2008} \shortcite{apsel_extended_2011} \shortcite{taghipour_generalized_2014} \shortcite{taghipour_completeness_2013-1} \shortcite{taghipour_completeness_2013} \shortcite{das_scaling_2016} \shortcite{taghipour_lifted_2012} \shortcite{taghipour_lifted_2013} \shortcite{ng_probabilistic_2008} \shortcite{ng_probabilistic_2009} \shortcite{takiyama_inference_2014} \shortcite{kisynski_lifted_2009} \shortcite{choi_efficient_2011} \shortcite{singla_lifted_2008} \shortcite{de_salvo_braz_anytime_2009} \shortcite{singla_approximate_2010} \shortcite{singla_approximate_2014} \shortcite{gogate_probabilistic_2016} \shortcite{gogate_advances_2012} \shortcite{van_den_broeck_lifted_2011} \shortcite{van_den_broeck_conditioning_2012} \shortcite{van_den_broeck_completeness_2011} \shortcite{van_den_broeck_skolemization_2014} \shortcite{meert_lifted_2014} \shortcite{beame_symmetric_2015} \shortcite{vlasselaer_knowledge_2016} \shortcite{bui_exact_2012} \shortcite{gogate_exploiting_2010} \shortcite{choi_lifted_2012-2} \shortcite{jha_lifted_2010} \shortcite{poole_towards_2011} \shortcite{kazemi_elimination_2014} \shortcite{kazemi_new_2016} \shortcite{kazemi_domain_2017} \shortcite{kiddon_leveraging_2010} \shortcite{kiddon_coarse--fine_2011} \shortcite{poon_general_2008} \shortcite{sarkhel_lifting_2013} \shortcite{sarkhel_lifted_2014} \shortcite{venugopal_just_2015} \shortcite{mittal_new_2014}  \shortcite{domingos_tractable_2012} \shortcite{dalvi_computing_2010} \shortcite{dalvi_efficient_2007} \shortcite{dylla_top-k_2013} \shortcite{jha_probabilistic_2012} \\ 
 Bottom-up LI  & \shortcite{kersting_informed_2010}  \shortcite{jaimovich_template_2007-1} \shortcite{kersting_counting_2009} \shortcite{ahmadi_mapreduce_2013} \shortcite{ahmadi_exploiting_2013} \shortcite{venugopal_non-parametric_2016} \shortcite{venugopal_scaling-up_2014} \shortcite{venugopal_lifting_2012} \shortcite{venugopal_evidence-based_2014} \shortcite{van_den_broeck_lifted_2012} \shortcite{hadiji_reduce_2013} \shortcite{sen_exploiting_2008} \shortcite{sen_bisimulation-based_2009} \shortcite{bui_automorphism_2013} \shortcite{bui_lifted_2014} \shortcite{niepert_markov_2012} \shortcite{anand_contextual_2016} \shortcite{van_den_broeck_lifted_2015} \shortcite{niepert_symmetry-aware_2013} \shortcite{mladenov_efficient_2014} \shortcite{apsel_lifting_2014} \shortcite{mladenov_lifted_2012}  \shortcite{mladenov_lifted_2013} \shortcite{mladenov_lifted_2014} \shortcite{van_den_broeck_complexity_2013}  \shortcite{nath_efficient_2010} \shortcite{nath_efficient_2010-1} \shortcite{ahmadi_lifted_2010} \shortcite{hadiji_efficient_2011} \shortcite{ahmadi_lifted_2011} \shortcite{geier_approximate_2011} \\ 
 Continuous Inference & \shortcite{belle_probabilistic_2015} \shortcite{belle_hashing-based_2015} \shortcite{belle_component_2016} \shortcite{sanner_symbolic_2012-1} \shortcite{shenoy_inference_2011}\\
  Logical Particle Filter & \shortcite{zettlemoyer_logical_2008} \\ 
 Relational Particle Filter & \shortcite{nitti_particle_2013} \shortcite{nitti_probabilistic_2016} \shortcite{nitti_relational_2014} \\ 
  Relational Kalman Filter & \shortcite{choi_lifted_2010} \shortcite{choi_lifted_2011} \shortcite{choi_learning_2015} \\ 
  Data Association & \shortcite{schumitsch_information-form_2005} \shortcite{huang_efficient_2009} \shortcite{huang_fourier_2009} \shortcite{huang_exploiting_2009} \shortcite{jagabathula_inferring_2011} \shortcite{kondor_multi-object_2007} \shortcite{jiang_fourier-information_2011} \shortcite{baum_association-free_2010} \shortcite{baum_using_2011} \shortcite{baum_kernel-sme_2013} \shortcite{baum_optimal_2012} \shortcite{baum_mmospa-based_2014} \shortcite{hanebeck_association-free_2015} \shortcite{leven_multiple_2004} \shortcite{leven_unscented_2009} \shortcite{mahler_multitarget_2003} \\ 
  Prob. Multiset Rewriting & \shortcite{barbuti_maximally_2011} \shortcite{barbuti_probabilistic_2012} \shortcite{krishnamurthy_biologically_2004} \shortcite{warnke_syntax_2015} \shortcite{bistarelli_representing_2003-1} \shortcite{oury_multi-level_2013} \shortcite{maus_rule-based_2011} \\ 
   \bottomrule 
   \end{longtable}

%% file: Input/07_AppendixD.tex

\section{List of Abbreviations}

\centering
\begin{tabular}{ll}
\toprule
Abbreviation & Explanation\\
\midrule
BP & Belief propagation \\
C-FOVE & Counting first-order variable elimination\\
DBN & Dynamic Bayesian network\\
FOVE & First-order variable elimination\\
LBP & Lifted belief propagation \\
LI & Lifted inference \\
LP & Linear program \\
LPF & Logical particle filter\\
MAP & Maximum-a-posteriori\\
MCMC & Markov chain Monte Carlo\\
MDP & Markov decision process\\
MLN & Markov logic network\\
MRS & Multiset rewriting system\\
RC & Recursive conditioning\\
RV & Random variable\\
VE & Variable elimination\\
WFOMC & Weighted first-order model counting\\
\bottomrule
\end{tabular}